\documentclass{elsarticle}
\usepackage{tabularray}
\usepackage{amssymb}
\usepackage{xcolor}
\usepackage{amsthm}
\usepackage{comment}
\usepackage{booktabs}
\usepackage[most]{tcolorbox}
\usepackage{enumitem}
\usepackage{subcaption}
\usepackage{listings}
\lstdefinelanguage{json}{
    morestring=[b]",
    morekeywords={true,false,null},
    sensitive=true
}
\usepackage{xcolor}
\usepackage{anyfontsize}
\usepackage{amsfonts}
\usepackage{algorithm}
\usepackage{algpseudocode}
\usepackage{amsmath}
\usepackage{dsfont}
\usepackage{url}
\usepackage{bbm}

\usepackage{graphicx}
\usepackage{subcaption}
\usepackage{multirow}
\usepackage{setspace}
\usepackage{makecell}
\usepackage{tcolorbox}
\usepackage{amsmath} 
\usepackage[framemethod=tikz]{mdframed}
\newenvironment{promptbox}[1]{%
  \begin{mdframed}[
    frametitle={#1},
    frametitlebackgroundcolor=gray!70,
    frametitlefont=\color{white}\bfseries\footnotesize,
    frametitlealignment=\raggedright,
    frametitleaboveskip=5pt,
    frametitlebelowskip=5pt,
    frametitlerule=false,
    linecolor=gray!70,
    backgroundcolor=white,
    linewidth=0.7pt,
    roundcorner=3pt,
    innerleftmargin=8pt,
    innerrightmargin=8pt,
    innertopmargin=7pt,
    innerbottommargin=7pt,
    skipabove=8pt,
    skipbelow=12pt
  ]%
  \footnotesize
}{%
  \end{mdframed}
}

\newcommand{\promptheading}[1]{%
  \par\medskip
  \noindent\textbf{#1}\par
  \smallskip
}
\newcounter{ex}

\biboptions{numbers,sort&compress}

\journal{JBI}

\begin{document}

\begin{frontmatter}

%% Title, authors and addresses

%% use the tnoteref command within \title for footnotes;
%% use the tnotetext command for theassociated footnote;
%% use the fnref command within \author or \address for footnotes;
%% use the fntext command for theassociated footnote;
%% use the corref command within \author for corresponding author footnotes;
%% use the cortext command for theassociated footnote;
%% use the ead command for the email address,
%% and the form \ead[url] for the home page:
%% \title{Title\tnoteref{label1}}
%% \tnotetext[label1]{}
%% \author{Name\corref{cor1}\fnref{label2}}
%% \ead{email address}
%% \ead[url]{home page}
%% \fntext[label2]{}
%% \cortext[cor1]{}
%% \affiliation{organization={},
%%             addressline={},
%%             city={},
%%             postcode={},
%%             state={},
%%             country={}}
%% \fntext[label3]{}

\title{ANCHOR-RE: An Agentic Biomedical Relation Extraction Framework with Retrieval Augmentation and Error-Driven Verification}

%% use optional labels to link authors explicitly to addresses:
% \author[label1,label2]{}
%% \affiliation[label1]{organization={},
%%             addressline={},
%%             city={},
%%             postcode={},
%%             state={},
%%             country={}}
%%
%% \affiliation[label2]{organization={},
%%             addressline={},
%%             city={},
%%             postcode={},
%%             state={},
%%             country={}}

\author[inst1]{Shufan Ming}

\affiliation[inst1]{organization={School of Information Sciences, University of Illinois Urbana-Champaign},
            addressline={501 E Daniel St.},
            city={Champaign},
            postcode={61820},
            state={IL},
            country={USA}}
\author[inst1]{Yikun Han}
\author[inst1]{Gibong Hong}
\author[inst2]{Rui Zhang}
\author[inst1]{Halil Kilicoglu}
\affiliation[inst2]{organization={Division of Computational Health Sciences, Department of Surgery, University of Minnesota},
            addressline={516 Delaware St SE},
            city={Minneapolis},
            postcode={55455},
            state={MN},
            country={USA}}

\begin{abstract}
\textbf{Objective.} Biomedical relation extraction (BioRE) extracts structured knowledge from biomedical literature for applications such as knowledge graph construction and hypothesis generation. In real-world settings, BioRE systems must distinguish true relations from numerous candidate entity pairs, most of which are unrelated. Traditional symbolic systems, such as SemRep, provide high precision and interpretability but have limited recall under linguistic variability and complex discourse structures. In contrast, large language models (LLMs) offer strong contextual reasoning capabilities but can produce false-positive relation predictions. This study aims to develop an agentic framework that augments LLM inference with retrieved demonstrations, external biomedical knowledge, and error-driven verification to improve BioRE reliability.\\
\textbf{Methods.} We developed \textsc{ANCHOR-RE}, a framework that integrates retrieved demonstrations, ontology-guided semantic constraints, external knowledge grounding, and data-driven verification rules into LLM inference. We evaluated it on three BioRE benchmarks (SemRepGS, DDI, and ChemProt) using both proprietary and open-weight LLMs. To assess generalizability beyond benchmark datasets while reducing potential evaluation bias from LLM pretraining contamination, we conducted a post-cutoff temporal precision evaluation using 100 biomedical articles published in 2026.\\
\textbf{Results.} With the proprietary LLM backbone, \textsc{ANCHOR-RE} outperformed direct prompting, improving micro-F$_1$ from 0.654 to 0.676 on SemRepGS, from 0.769 to 0.881 on DDI, and from 0.937 to 0.941 on ChemProt, outperforming the previously reported inference-only methods on DDI and ChemProt. Performance gains were also observed with the open-weight LLM backbone, while additional experiments demonstrated cross-family model compatibility and sensitivity to the quality of external knowledge. On the post-cutoff set, manual assessment of 500 randomly sampled positive predictions yielded a precision of 69\% on recently published biomedical literature.\\
\textbf{Conclusion.} Augmenting LLM inference with contextual evidence and error-driven verification can improve the reliability of BioRE without task-specific fine-tuning. Results across multiple benchmarks, model families, and post-cutoff literature support \textsc{ANCHOR-RE} as a practical training-free approach to biomedical literature mining. Data and code are available at \url{https://github.com/COMBINI-Hub/Multi_Agent_Relation_Extraction}.
\end{abstract}

%%Graphical abstract
% \begin{graphicalabstract}
% \includegraphics{grabs}
% \end{graphicalabstract}

%%Research highlights
% \begin{highlights}
% \item Research highlight 1
% \item Research highlight 2
% \end{highlights}

\begin{keyword}
%% keywords here, in the form: keyword \sep keyword
Biomedical relation extraction \sep Large language models \sep Biomedical literature \sep SemRep
%% PACS codes here, in the form: \PACS code \sep code
% \PACS 0000 \sep 1111
%% MSC codes here, in the form: \MSC code \sep code
%% or \MSC[2008] code \sep code (2000 is the default)
% \MSC 0000 \sep 1111
\end{keyword}

\end{frontmatter}

%% \linenumbers

%% main text
\section{Introduction}
\label{sec:intro}
Biomedical relation extraction (BioRE) automatically identifies semantic relationships between biomedical entities, transforming unstructured biomedical literature into structured knowledge for downstream applications such as biomedical knowledge graph construction \cite{zhang2021drug, xu2025pubmed}, literature-based discovery \cite{henry2017literature}, and clinical decision support \cite{eguia2024clinical}. Previous BioRE approaches can generally be categorized into rule-based systems and supervised transformer-based models, each with distinct strengths and limitations. Rule-based systems such as SemRep leverage handcrafted lexical triggers and curated semantic type constraints to achieve high precision and provide interpretable predictions \citep{kilicoglu2020broad}, but have difficulty in generalizing across diverse linguistic expressions and complex discourse structures. In contrast, transformer-based models learn rich contextual representations and achieve competitive performance on benchmark datasets \citep{jullien2024semeval}. However, these models typically rely on task-specific fine-tuning using large annotated corpora, making adaptation to new biomedical domains costly and limiting applicability where labeled data are scarce \citep{meesawad2025enhancing, delmas2024relation}.

Generative large language models (LLMs) have demonstrated strong zero-shot and few-shot capabilities across a wide range of biomedical NLP tasks. However, even state-of-the-art models remain prone to hallucinations and factual inconsistencies, limiting their reliability in high-stakes biomedical applications \citep{frasca2024explainable}. In BioRE, these failures often manifest as systematic over-prediction, where entity co-occurrences are incorrectly inferred as meaningful relations \citep{yang2025depth}. Such limitations motivate inference-time approaches that augment LLM reasoning with relevant evidence and explicit verification to reduce unsupported relation predictions.

ReOnto \citep{jain2023reonto} incorporates ontology-derived relational paths into BioBERT representations, providing explicit relational cues to improve predictions when textual evidence is weak or ambiguous. Similarly, \citet{olasunkanmi2025relate} integrate ontology constraints into an LLM-based framework by restricting candidate predicates before contextual reasoning. Despite their promising performance, these approaches generally require either task-specific model training or specialized multi-stage reasoning pipelines. Lightweight inference-time methods that inject constraints without additional model training remain largely underexplored.

To address this gap, we propose \textsc{ANCHOR-RE}, an agentic framework that augments relation prediction with contextual evidence, including retrieved demonstrations and KB evidence, and applies error-driven verification to improve LLM-based BioRE. It consists of two modules: a \textsc{Decider} that generates candidate relation predictions, and a \textsc{Verifier} that verifies these predictions using learned error patterns derived from the training corpus. Entity semantic type constraints further restrict candidate relations based on subject-object semantic type compatibility, reducing biologically implausible predictions. 

We evaluate \textsc{ANCHOR-RE} on three BioRE benchmarks spanning diverse relation schemas using both proprietary and open-weight LLMs. To further assess robustness beyond standard benchmark settings, we additionally conduct a post-cutoff temporal evaluation on biomedical articles published after the knowledge cutoff of the underlying LLM. We assess the contribution of each framework component through ablation studies and further investigate its effectiveness in reducing hallucinated false-positive predictions. Experimental results demonstrate that augmenting LLM inference with demonstration retrieval, external evidence, and error-driven verification can improve performance over direct prompting, with gains varying across benchmarks and model backbones.

The main contributions of this study are as follows: 
\begin{itemize}
\item We propose \textsc{ANCHOR-RE}, a training-free agentic framework that augments LLM-based BioRE with different sources of evidence, including retrieved demonstrations, KB evidence, and semantic type constraints.
\item We introduce a reusable verification module (\textsc{Verifier}) that applies natural-language criteria learned from recurring error patterns as soft constraints to reduce false positive BioRE predictions at inference time.
\item We demonstrate the robustness and generalizability of ANCHOR-RE through comprehensive evaluation on three BioRE benchmarks, post-cutoff temporal evaluation, and experiments with both proprietary and open-weight LLM backbones.
\end{itemize}

\begin{table}
\scriptsize
\begin{tabular}{|p{3cm}|p{8.5cm}|}
\hline
Problem &
BioRE is a fundamental task in mining structured knowledge from biomedical literature that supports downstream applications such as knowledge graph construction, literature-based discovery, and clinical decision support. However, current LLM-based approaches remain prone to hallucinations and often generate biologically unsupported false-positive predictions, limiting their reliability in real-world biomedical applications.
\\ \hline

What is already known &
Recent inference-time knowledge augmentation approaches incorporate structured biomedical knowledge into neural models to improve BioRE performance. However, these methods typically require task-specific model training or complex multi-stage reasoning pipelines, limiting their flexibility and scalability across diverse BioRE settings.
\\ \hline

What this paper adds &
We propose \textsc{ANCHOR-RE}, a training-free framework that integrates knowledge base evidence, semantic type constraints, and inference-time verification to improve LLM-based BioRE. Experiments on three benchmark datasets and a post-cutoff temporal evaluation demonstrate improved reliability, reduced hallucinated false-positive predictions, and generalizability without additional model training.
\\ \hline

Who would benefit &
Researchers and practitioners developing biomedical text mining systems, knowledge graphs, and AI-driven applications that leverage structured knowledge from the literature. 
\\ \hline
\end{tabular}
\caption{Statement of significance.}
\label{significance_statement}
\end{table}

\section{Related Work}

\subsection{LLMs for Biomedical Relation Extraction}

Generative LLMs (e.g., GPT, Qwen, LLaMA) have demonstrated strong zero-shot and few-shot capabilities for BioRE, substantially reducing the dependence on task-specific annotated corpora while achieving competitive performance on several BioRE benchmarks and, in some settings, approaching or surpassing supervised approaches \citep{brokman2025benchmark, gade2025benchmarking, li2023revisiting}. Recent inference-only methods have further improved BioRE through prompt engineering and retrieval augmentation.
Liu et al. proposed a self-prompting framework that automatically generates diverse synthetic examples that are subsequently used as in-context demonstrations for zero-shot relation extraction \citep{liu2024unleashing}. Retrieval-augmented generation (RAG) methods improve grounding by incorporating external biomedical evidence during inference. For example, BiomedRAG retrieves biomedical evidence and trains a relevance scorer to rank and filter candidate passages, ensuring that only the most informative content is provided to the LLM for reasoning and information extraction \citep{li2025biomedrag}. Despite these advances, most inference-only BioRE approaches rely primarily on prompt engineering or retrieved unstructured text and therefore provide limited mechanisms for enforcing explicit biomedical semantic constraints or verifying relation predictions.

\subsection{Inference-time Knowledge Augmentation and Verification for Biomedical Relation Extraction}
To improve the reliability of LLM-based BioRE, recent approaches have incorporated structured biomedical knowledge, such as ontologies, semantic type constraints, and symbolic rules, into the inference process. Quan et al.~\citep{quan2025peirce} introduce a symbolic validation module that filters predictions violating ontology-based constraints. Zhao et al.~\citep{zhao2025zero} incorporate entity synonym and hypernym knowledge into prompting, providing both positive and negative textual evidence to guide reasoning. Li et al.~\citep{li2024recall} propose a recall--retrieve--reason framework that leverages ontology knowledge to generate retrieval queries for candidate entity pairs, which are then used to retrieve in-context demonstrations for relation extraction.

Despite these advances, recent benchmarking studies show that LLM performance decreases as relation schemas become larger and semantically more complex, making BioRE particularly challenging in realistic open biomedical settings \citep{sanger2025knowledge}. Moreover, LLMs remain prone to overconfident and unsupported relation predictions \citep{de2025study}. Our work investigates an inference-only framework that integrates multiple grounding knowledge including knowledge-base evidence, semantic type constraints, and learned verification rules, directly into LLM reasoning to improve the reliability of relation prediction.

\section{Methods}

\subsection{Task Formulation}
The goal of the BioRE is to predict the relation label between a given pair of entities within a sentence\footnote{While document-level RE is also an important setting, we focus on sentence-level RE in this work.}. Formally, given an input sentence $s$ and a target entity pair $(e_1,e_2)$ together with their semantic types, the objective is to predict a relation label $\hat{r}\in\mathcal{R}$, where $\mathcal{R}$ denotes a predefined relation set, including a \textsc{norel} label that represents the absence of a valid relation.

Our baseline follows a standard zero-shot LLM classification setting. The baseline prompt contains only the task instruction, dataset-specific relation definitions, the input sentence, and the target entity pair (See~\ref{app:baseline_prompt}). \textsc{ANCHOR-RE} extends this baseline by augmenting LLM inference with three complementary sources of guidance: retrieved in-context demonstrations selected from the training corpus, structured evidence retrieved from external biomedical KBs, and verification rules learned offline from systematic false-positive predictions.

\subsection{Overview of ANCHOR-RE}

%Figure~\ref{fig:framework} provides an overview of \textsc{ANCHOR-RE}, which consists of an offline preparation stage followed by an inference stage. During the offline stage, the labeled training corpus is embedded into a vector database for demonstration retrieval. To construct the \textit{Error Pattern Knowledge Bank}, we first collect systematic false-positive predictions produced by the zero-shot baseline inference on the training corpus. These failure cases are subsequently analyzed through an additional LLM inference to derive a set of reusable verification rules. This offline procedure is performed once for each dataset and inference backbone LLM and does not require parameter updates or fine-tuning.

\begin{figure*}[htbp]
    \centering
    \includegraphics[width=\textwidth]{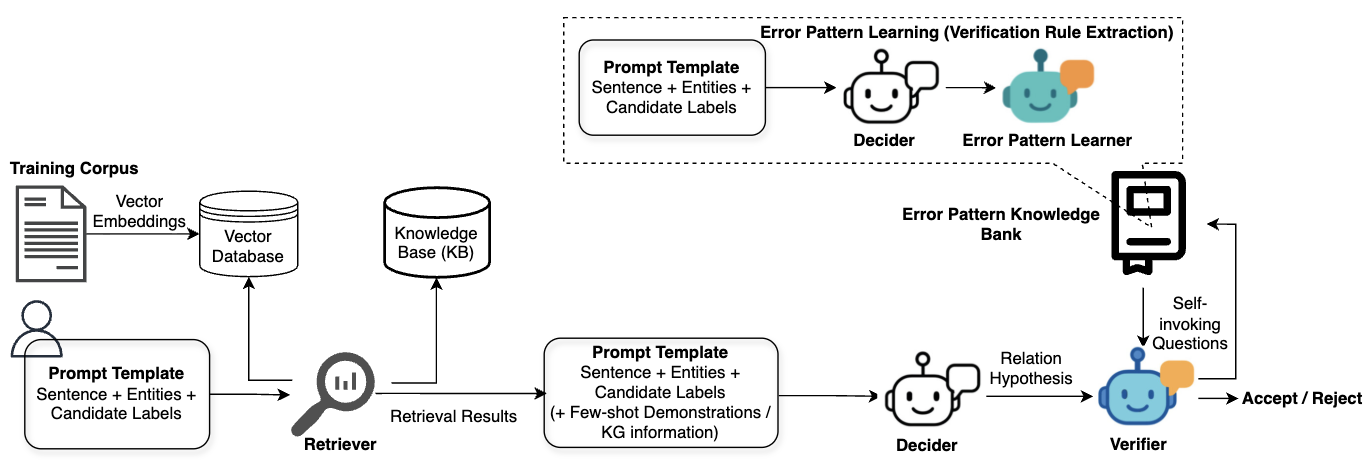}
    \vspace{-5pt}
    \caption{
    \small
    Overview of \textsc{ANCHOR-RE}.
    }
    \label{fig:framework}
    \vspace{-10pt}
\end{figure*}

Figure~\ref{fig:framework} provides an overview of \textsc{ANCHOR-RE}, which consists of an offline preparation stage and an inference stage. The offline stage constructs two resources that support subsequent inference: (1) a vector database containing embeddings of labeled training examples for retrieval of in-context demonstrations, and (2) an \textit{Error Pattern Knowledge Bank} that captures recurring false-positive patterns and corresponding verification rules. To construct the latter, we first perform zero-shot inference on the training corpus to identify systematic false-positive errors. These errors are then analyzed using an additional LLM-based inference module (\textsc{Error Pattern Learner}) to derive reusable error patterns and verification rules. This preparation procedure is performed once for each dataset and LLM backbone and does not require model parameter updates or fine-tuning.

During inference, \textsc{ANCHOR-RE} first generates a candidate relation prediction using the \textsc{Decider}, which leverages the input sentence, entity pair, retrieved demonstrations, and structured evidence from external KBs, with ontological constraints further restricting candidate relations where applicable. The candidate prediction is then examined by a verification module (\textsc{Verifier}) that evaluates it against the error patterns and rules in the \textit{Error Pattern Knowledge Bank} and produces the final relation label.

Formally, the overall prediction process is defined as

\[
\hat{r}
=
g\!\left(
f_{\theta}(s,e_1,e_2;
\mathcal{D}_{\mathrm{retr}},
K_{\mathrm{ext}}),
K_{\mathrm{err}}
\right),
\]

where $f_{\theta}$ represents the \textsc{Decider}, an LLM parameterized by $\theta$ which generates a candidate relation based on the input sentence $s$, entity pair $(e_1,e_2)$, retrieved demonstrations $\mathcal{D}_{\mathrm{retr}}$, and structured evidence from external KBs $K_{\mathrm{ext}}$. $K_{\mathrm{err}}$ denotes the \textit{Error Pattern Knowledge Bank} constructed during offline preparation, and $g(\cdot)$ represents the verification module that refines the candidate prediction using the learned verification rules to produce the final relation label.

The following subsections describe the three main modules of \textsc{ANCHOR-RE}: the \textsc{Decider}, which performs retrieval-augmented relation prediction; the offline \textsc{Error Pattern Learner}, which derives verification criteria from recurring false-positive errors; and the \textsc{Verifier}, which applies these criteria to verify predictions at inference time.

%Figure~\ref{fig:framework} provides an overview of \textsc{ANCHOR-RE}, which consists of an offline preparation stage and an inference stage. The offline stage constructs two resources that support inference: (1) a vector database containing embedded examples from the labeled training corpus for demonstration retrieval and (2) an \textit{Error Pattern Knowledge Bank} containing verification rules derived from recurring false-positive patterns. To construct the latter, we first perform zero-shot inference on the training corpus to identify erroneous predictions and then use an additional LLM-based analysis step to extract reusable error patterns and corresponding verification rules. This preparation is performed once for each dataset and backbone LLM and does not require parameter updates or model fine-tuning.

%Formally, the prediction process is defined as:
% \[
% \hat{r}
% =
% g\!\left(
% f_{\theta}(s,e_1,e_2;
% \mathcal{D}_{\mathrm{retr}},
% K_{\mathrm{ext}}),
% P_{\mathrm{err}}
% \right),
% \]

% where $f_{\theta}$ represents the \textsc{Decider}, which generates a candidate relation hypothesis;  $\mathcal{D}_{\mathrm{retr}}$ represents retrieved in-context demonstrations, $K_{\mathrm{ext}}$ denotes structured evidence retrieved from external knowledge bases, and $P_{\mathrm{err}}$ is the Error Pattern Knowledge Bank learned during the offline stage. The verification function $g(\cdot)$ evaluates the candidate prediction against the learned verification rules to produce the final relation label.

\subsection{Retrieval-Augmented Relation Prediction}
The \textsc{Decider} is the inference module for generating a candidate relation prediction for a given sentence and target entity pair. Inspired by symbolic systems such as SemRep, which identify relations using lexical triggers (e.g., verbs, nominalizations, and prepositional cues), we incorporate trigger-oriented guidance into the \textsc{Decider} prompt. Specifically, prompts include structured descriptions of common relational triggers, their lexical forms, typical syntactic patterns, and trigger–relation mappings. These cues act as soft linguistic constraints that highlight plausible relational signals while allowing the LLM to generalize beyond hand-crafted rules. 

In addition to the task instructions, the \textsc{Decider} is provided with two complementary sources of supporting evidence: (i) retrieved training instances used as in-context demonstrations ($\mathcal{D}_{\mathrm{retr}}$), and (ii) structured relational evidence retrieved from external biomedical KBs ($K_{\mathrm{ext}}$) (see~\ref{app:baseline_prompt}). Based on the augmented prompt, the \textsc{Decider} generates a structured prediction grounded in a trigger span copied from input sentence, a brief explanation, and the predicted relation label.

\subsubsection{Demonstration Selection}
For each training instance $x_i$, we construct a structured representation by serializing the sentence $s$, target entity pair $(e_1,e_2)$, and relation label $r$ into a templated sequence. By jointly embedding the relation label together with the entity names and semantic types, the resulting representation captures relation-specific semantics and semantic-type compatibility beyond surface-level sentence similarity. This sequence is then encoded using a PubMedBERT-based encoder to obtain a representation $h_{x_i} \in \mathbb{R}^d$. 

At inference time, given a test instance $x$, we consider each candidate relation label $r \in \mathcal{R}(x)$ independently and retrieve demonstrations conditioned on this label. Retrieved examples are ranked by prioritizing exact subject--object semantic-type matches, followed by semantic-group matches and then the remaining examples; within each category, examples are ordered by cosine similarity between the dense representations of the test instance and each retrieved example.
% Retrieved examples are ranked using the similarity score: 
% \[
% \text{score}_r(x, x_i) 
% = \alpha \cdot \mathrm{sim}(h_x, h_{x_i}) 
% + \beta \cdot \mathbbm{1}\!\left[\mathrm{type}(x) = \mathrm{type}(x_i)\right],
% \] where $\mathrm{sim}$ denotes cosine similarity, and the indicator function prioritizes examples whose subject–object semantic type pair matches that of the test instance. 
We incorporate contrastive in-context demonstrations by retrieving a balanced set of positive examples (instances labeled with a relation in the candidate set $\mathcal{R}$) and negative examples (instances labeled as \textsc{norel}). The number of retrieved demonstrations, $k$, is tuned on the validation set.

\subsubsection{Structured Relational Evidence Retrieval}
In addition to retrieved demonstrations, we query KBs conditioned on the target entity pair $(e_1, e_2)$. We leverage dataset-specific sources, including the UMLS Metathesaurus \citep{bodenreider2004unified} for SemRepGS, DrugBank~\citep{wishart2008drugbank} for DDI~\citep{herrero2013ddi}, and CTD~\citep{mattingly2006comparative} for ChemProt~\citep{krallinger2017overview}. For SemRepGS, we query concepts and relations from controlled biomedical vocabularies integrated in UMLS (e.g., SNOMED CT, MeSH, RxNorm, and NCI Thesaurus), and explicitly incorporate semantic relations compatible with the UMLS Semantic Network.
% (e.g., \textit{sulfonylureas}-\textsc{TREATS}-\textit{type 2 diabetes}). 

The retrieved knowledge is organized into evidence blocks and appended to the \textsc{Decider} prompt to provide KB-level grounding. Depending on the dataset and available KB information, these blocks can include direct and neighboring concept relations for SemRepGS, source-provided fields such as drug descriptions and neighboring interactions from DrugBank for DDI, and entity-level information and neighboring chemical--gene relations from CTD for ChemProt. No additional LLM-based processing is used to generate these evidence blocks. Examples of retrieved external knowledge are provided in~\ref{app:kg-examples}.

\subsection{Learning Verification Rules from False Positives}
\label{constructionmethod}
We perform zero-shot inference with the frozen \textsc{Decider} on a stratified subset of the training corpus without retrieval augmentation. For each relation label $r$, we construct a failure pool $\mathcal{F}_r$ consisting of instances incorrectly predicted as $r$ despite having no valid relation:
\[
\mathcal{F}_r = \{ x \mid \hat{y}(x)=r \;\wedge\; y(x)=\textsc{norel} \}.
\]
This pool captures examples of systematic over-predictions and serves as the basis for learning verification rules.

\textsc{Error Pattern Learner}, an offline LLM-based module, induces verification rules from systematic false-positive predictions. The LLM is prompted with examples from the failure pool, together with true-positive examples and previously learned verification rules, to induce abstract verification rules represented as structured questions (see \ref{learner} for the full prompt). During each refinement iteration, the LLM is instructed to revise, merge, or introduce new rules when existing rules are overly broad, overly specific, redundant, or likely to reject valid positive predictions.
Through iterative refinement, this process generalizes from instance-level errors to label-level soft constraints, forming a per-label \textit{Error Pattern Knowledge Bank} that captures systematic over-prediction biases of the \textsc{Decider}. After each refinement iteration, the resulting \textit{Error Pattern Knowledge Bank} is evaluated on a held-out validation set, and the version achieving the highest validation micro-F$_1$ is selected for inference.

\subsection{Verification}
During inference, the \textsc{Verifier} is invoked only when the \textsc{Decider} predicts a positive relation. It retrieves the verification rules for the predicted relation from the precomputed \textit{Error Pattern Knowledge Bank} and evaluates whether any rule applies to the current prediction. If any verification rule is triggered, the prediction is rejected and relabeled as \textsc{norel}; otherwise, the original prediction is retained. The full verification prompt is provided in~\ref{Verifier}.

The overall procedure for error pattern learning is summarized in Algorithm~\ref{alg:pattern_distillation}. In the algorithm, \textsc{ExtractVerificationRules} extracts candidate verification rules from batches of failure cases and uses them to construct the \textit{Error Pattern Knowledge Bank}. \textsc{ValidationSelect} evaluates the resulting pattern bank on the validation set, and retains the version that achieves the highest validation micro-F$_1$. During inference, the \textsc{Verifier} applies the relevant soft constraints in the \textit{Error Pattern Knowledge Bank}.
% \textcolor{red}{Unclear to me what \textsc{ValidationSelect} is in the algorithm, also why return P=0 at the end of the algorithm?}

\begin{algorithm}[h]
\caption{Error Pattern Learning for Verification}
\small
\label{alg:pattern_distillation}
\begin{algorithmic}[1]
\Require 
Training set $\mathcal{D}_{\mathrm{train}}$ with gold labels $y(x)$;
validation set $\mathcal{D}_{\mathrm{val}}$;
relation set $\mathcal{R}$ with semantic definitions $\{\delta_r\}$; 
frozen \textsc{Decider} model $f_\theta$;
batch size $B$;
maximum refinement iterations $T$
\Ensure Validation-selected Error Pattern Knowledge Bank $\mathcal{P}$
\State $\mathcal{P} \leftarrow \emptyset$
\State \textbf{/* Stage 1: Probe model decision behavior */}
\State $\mathcal{S} \leftarrow \Call{StratifiedSample}{\mathcal{D}_{\mathrm{train}}}$
\For{each instance $x \in \mathcal{S}$}
    \State $\hat{y}(x) \leftarrow f_\theta(x)$
\EndFor
\For{each relation $r \in \mathcal{R}$}
    \State $\mathcal{F}_r \leftarrow \{x \in \mathcal{S} \mid \hat{y}(x)=r \land y(x)=\textsc{norel}\}$  
    \State $\mathcal{T}_r \leftarrow \{x \in \mathcal{S} \mid \hat{y}(x)=r \land y(x)=r\}$  
    \State $\mathcal{P}_r \leftarrow \emptyset$
    \State \textbf{/* Stage 2: Iterative rule learning */}
    \For{$t = 1,\dots,T$}
        \State $\mathcal{B} \leftarrow \Call{NextBatch}{\mathcal{F}_r, B}$
        \State $\Delta_r \leftarrow 
        \Call{ExtractVerificationRules}{
            \mathcal{B}, 
            \mathcal{T}_r, 
            \delta_r, 
            \mathcal{P}_r
        }$
        \State $\mathcal{P}_r \leftarrow 
        \Call{ReviseAndUpdate}{
            \mathcal{P}_r, 
            \Delta_r
        }$
        \If{$\Call{NoNewPatterns}{\Delta_r}$}
            \State \textbf{break}
        \EndIf
    \EndFor
    \State $\mathcal{P} \leftarrow \mathcal{P} \cup \mathcal{P}_r$
\EndFor
\State \textbf{/* Stage 3: Validation-set selection */}
\State $\mathcal{P} \leftarrow \Call{ValidationSelect}{\mathcal{P}, \mathcal{D}_{\mathrm{val}}}$
\State \Return $\mathcal{P}$
\end{algorithmic}
\end{algorithm}

\subsection{Datasets}

We evaluate our framework on three benchmark BioRE datasets, each formulated as a sentence-level multi-class relation classification task over predefined entity pairs. Dataset statistics are provided in~\ref{app:data-stats}.

\paragraph{\textbf{SemRepGS}} 
We evaluate our framework on SemRepGS \citep{ming2024enhancing} adapted from the original SemRep gold standard dataset \citep{kilicoglu2011constructing}. The dataset consists of sentence-level subject–predicate–object annotations and includes 23 relation types (including \textsc{norel}). For each instance, the candidate relation set is pruned using UMLS Semantic Network constraints, retaining only relations compatible with the semantic types of the target subject and object. For example, \textsc{treats} is only considered when the subject is a therapeutic agent or procedure and the object is a disorder. On average, 3 candidate relations per entity pair remain after filtering.
This pruning step reduces implausible relation candidates prior to inference and reflects the ontology-driven constraints used in prior rule-based relation extraction systems~\citep{kilicoglu2020broad}. 

\paragraph{\textbf{DDI}} We evaluate on the Drug–Drug Interaction (DDI) extraction dataset released in the SemEval-2013 shared task \citep{herrero2013ddi}. The dataset contains annotations of pharmacological interactions between drug entity pairs. Since all entities are drugs, semantic type-based constraints are not required for this dataset.

\paragraph{\textbf{ChemProt}}
ChemProt contains relations between chemical and protein entities. Following prior work \citep{krallinger2017overview,zhou2024leap}, we use the official data splits and retain only candidate pairs annotated with one of the five relation categories evaluated in the benchmark: CPR:3 (activation/upregulation), CPR:4 (inhibition/downregulation), CPR:5 (agonist), CPR:6 (antagonist), and CPR:9 (substrate or product-of). Since the \textsc{norel} class is not considered, we do not apply the \textsc{Verifier} module for this dataset.

\subsection{Experimental Setup}
\paragraph{\textbf{LLM settings}} We use GPT-5 (\texttt{gpt-5-mini-2025-08-07}) as the offline \textsc{Error Pattern Learner} in every reported configuration, selected for its balance between reasoning capability and inference efficiency. For the online \textsc{Decider} and \textsc{Verifier}, we evaluate GPT-5 and the open-weight model Qwen3 (\texttt{Qwen3-32B}) using the same prompting and inference pipeline. We further examine cross-family verification on SemRepGS by exchanging the GPT-5 and Qwen3 models used as the \textsc{Verifier}, while keeping the learned patterns and corresponding \textsc{Decider} predictions fixed. 

GPT-5 inference was conducted through the OpenAI API. For Qwen3, we used greedy decoding (temperature $=0$) with a fixed seed of 42. Inference was performed using vLLM on a single NVIDIA H200 GPU (141 GB VRAM), with 10 CPU cores and 32 GB host memory. Inference latency and API cost are reported in~\ref{latencycost}. 

\paragraph{\textbf{Pattern learning and dataset-specific settings}} The \textit{Error Pattern Knowledge Bank} is constructed using a batch size of $B=15$ and a maximum of $T=6$ refinement iterations. The pattern bank achieving the highest validation micro-F$_1$ is selected for inference. 

We divided the official DDI training set into error-pattern-learning (80\%) and validation (20\%) subsets using stratified sampling because the original dataset release does not provide a separate validation set. 

For in-context demonstration retrieval, $k$ is selected based on validation performance (SemRepGS: 3, DDI: 4, ChemProt: 2). SemRepGS and DDI use $k$ positive and $k$ negative examples, whereas ChemProt uses only positive examples. 

KB evidence is dataset-specific: SemRepGS used the 2025AB release of UMLS, which may include direct relations between the target entities that overlaps with the label set. DDI information was constructed from DrugBank version 5.1.13, while ChemProt used the November 2025 CTD snapshot (see \ref{app:kg-examples}). 

\paragraph{\textbf{Evaluated configurations}} For SemRepGS and DDI, we evaluated:
\begin{enumerate}[nosep]
\item \textsc{Baseline}: direct prompting;
\item + Retrieval: retrieved demonstrations without KB context;
\item + Retrieval + \textsc{Verifier}: verification of the + Retrieval predictions;
\item + KB: KB context without retrieved demonstrations;
\item + KB + Retrieval: both sources of context; and
\item + KB + Retrieval + \textsc{Verifier}: verification of the + KB + Retrieval predictions.
\end{enumerate}
For ChemProt, we evaluated the four configurations without the \textsc{Verifier} because the evaluation includes only positive relation categories.

\subsection{Evaluation}
We evaluate predictions at the entity-pair level using dataset-specific protocols. For SemRepGS, following prior work, \textsc{norel} is excluded when computing evaluation metrics because it denotes the absence of a relation rather than a positive semantic relation type. We therefore report micro- and macro-averaged precision, recall, and F$_1$ over the positive relation types. For DDI, we evaluate all relation classes, including \textsc{norel}. For ChemProt, we evaluate the five retained positive classes. Because both are single-label multiclass tasks, micro-precision, micro-recall, and micro-F$_1$ are identical. We therefore report only micro-F$_1$ together with macro-averaged precision, recall, and F$_1$.

\subsubsection{Post-cutoff Temporal Precision Evaluation}
LLMs may have been exposed to benchmark datasets during pretraining, which can lead to overly optimistic performance estimates due to data contamination \citep{sainz2023nlp}. To mitigate this concern and assess generalization beyond the pretraining period, we construct a temporal out-of-distribution evaluation set using 100 articles published in 2026. Titles and abstracts are processed with PubTator3 \citep{wei2024pubtator} to obtain biomedical entity annotations. The recognized entities are mapped to UMLS semantic types using rule-based heuristics (e.g., \textit{Disease} entities are mapped to \textit{Disease or Syndrome}). Candidate relation filtering based on UMLS Semantic Network constraints is then applied to reduce implausible relation types. The resulting dataset contains 3,890 candidate instances, with an average of 5 candidate relation labels per instance after filtering. We manually evaluated 500 randomly sampled positive predictions to estimate precision. The assessment was performed by one author (HK), who participated in SemRepGS annotation. This evaluation is precision-only, because ground truth relations are not available for this set.

\begin{table*}[t]
\centering
\small
\renewcommand{\arraystretch}{1.20}
% \makebox[\linewidth][c]{%
\resizebox{1.0\linewidth}{!}{%
\begin{tabular}{lcccccc}
\toprule
\multirow{2}{*}{Configuration} &
\multicolumn{3}{c}{Micro} &
\multicolumn{3}{c}{Macro} \\
\cmidrule(lr){2-4}
\cmidrule(lr){5-7}
&
Precision &
Recall &
F$_1$ &
Precision &
Recall &
F$_1$ \\
\midrule

\multicolumn{7}{l}{\textit{\textbf{SemRepGS}}} \\
\midrule

\textsc{Baseline}
& \makecell{0.534 \\ {[0.509, 0.560]}}
& \makecell{0.844 \\ {[0.821, 0.866]}}
& \makecell{0.654 \\ {[0.630, 0.677]}}
& \makecell{0.473 \\ {[0.439, 0.507]}}
& \makecell{0.793 \\ {[0.757, 0.828]}}
& \makecell{0.573 \\ {[0.535, 0.600]}} \\

+ Retrieval
& \makecell{0.553$^{***}$ \\ {[0.528, 0.578]}}
& \makecell{\textbf{0.858} \\ {[0.837, 0.878]}}
& \makecell{0.672$^{***}$ \\ {[0.651, 0.695]}}
& \makecell{0.495$^{**}$ \\ {[0.457, 0.529]}}
& \makecell{\textbf{0.818}$^{**}$ \\ {[0.785, 0.848]}}
& \makecell{0.595$^{**}$ \\ {[0.556, 0.624]}} \\

+ Retrieval w/\textsc{Verifier}
& \makecell{0.575$^{***\dagger\dagger\dagger}$ \\
  {[0.550, 0.601]}}
& \makecell{0.815$^{***\dagger\dagger\dagger}$ \\
  {[0.790, 0.838]}}
& \makecell{0.674$^{***}$ \\
  {[0.652, 0.696]}}
& \makecell{\textbf{0.516}$^{***\dagger\dagger\dagger}$ \\
  {[0.477, 0.554]}}
& \makecell{0.791$^{\dagger\dagger\dagger}$ \\
  {[0.757, 0.823]}}
& \makecell{\textbf{0.601}$^{***}$ \\
  {[0.561, 0.630]}} \\

\addlinespace[0.6em]

+ KB
& \makecell{0.534 \\ {[0.510, 0.560]}}
& \makecell{0.846 \\ {[0.823, 0.869]}}
& \makecell{0.655 \\ {[0.632, 0.679]}}
& \makecell{0.470 \\ {[0.435, 0.503]}}
& \makecell{0.790 \\ {[0.754, 0.824]}}
& \makecell{0.569 \\ {[0.530, 0.596]}} \\

+ KB + Retrieval
& \makecell{0.541 \\ {[0.516, 0.567]}}
& \makecell{0.853 \\ {[0.831, 0.874]}}
& \makecell{0.662 \\ {[0.640, 0.684]}}
& \makecell{0.480 \\ {[0.446, 0.514]}}
& \makecell{0.791 \\ {[0.754, 0.828]}}
& \makecell{0.579 \\ {[0.541, 0.607]}} \\

+ KB + Retrieval w/\textsc{Verifier}
& \makecell{\textbf{0.576}$^{***\dagger\dagger\dagger}$ \\
  {[0.551, 0.603]}}
& \makecell{0.816$^{***\dagger\dagger\dagger}$ \\
  {[0.792, 0.839]}}
& \makecell{\textbf{0.676}$^{***\dagger\dagger\dagger}$ \\
  {[0.654, 0.698]}}
& \makecell{0.512$^{***\dagger\dagger\dagger}$ \\
  {[0.475, 0.549]}}
& \makecell{0.766$^{*\dagger\dagger\dagger}$ \\
  {[0.728, 0.803]}}
& \makecell{0.597$^{**\dagger\dagger\dagger}$ \\
  {[0.557, 0.626]}} \\

\bottomrule
\end{tabular}%
}
% \vspace{1mm}
% \begin{minipage}{0.98\linewidth}
% \scriptsize
% Numbers in brackets denote 95\% confidence intervals obtained from bootstrap evaluation ($n=1{,}000$). Statistical significance is calculated using paired bootstrap tests against \textsc{Baseline} ($^{*}p<0.1$, $^{**}p<0.05$, and $^{***}p<0.01$) and for configurations with the \textsc{Verifier},  against the corresponding configuration without the \textsc{Verifier} ($^{\dagger\dagger\dagger}p<0.01$).\end{minipage}
\caption{Evaluation results on the SemRepGS test set using GPT-5 as the backbone LLM. The \textsc{norel} class was excluded when computing micro- and macro-averaged metrics. Numbers in brackets denote 95\% confidence intervals obtained from bootstrap evaluation ($n=1{,}000$). Statistical significance is calculated using paired bootstrap tests against \textsc{Baseline} ($^{*}p<0.1$, $^{**}p<0.05$, and $^{***}p<0.01$) and for configurations with the \textsc{Verifier}, against the corresponding configuration without the \textsc{Verifier} ($^{\dagger\dagger\dagger}p<0.01$).}
\label{tab:test_results_semrep}
\end{table*}

\section{Results}
\subsection{\textsc{ANCHOR-RE} performance on three benchmarks}
\paragraph{\textbf{SemRepGS}} We first analyze how different components of \textsc{ANCHOR-RE} contribute to performance on the SemRepGS benchmark (Table~\ref{tab:test_results_semrep}). Demonstration retrieval yields clear gains, increasing micro-F$_1$ from 0.654 to 0.672 ($p<0.01$) and macro-F$_1$ from 0.573 to 0.595 ($p<0.05$). Incorporating KB context alone yields no significant improvement over the \textsc{Baseline}, with micro-F$_1$ increasing marginally from 0.654 to 0.655 and macro-F$_1$ decreasing slightly from 0.573 to 0.569. Adding KB context to retrieval does not provide an additional benefit: + KB + Retrieval achieves a micro-F$_1$ of 0.662 and a macro-F$_1$ of 0.579, both lower than the corresponding + Retrieval results.

The + KB + Retrieval + \textsc{Verifier} configuration achieves the highest micro-F$_1$ of 0.676, whereas + Retrieval + \textsc{Verifier} achieves the highest macro-F$_1$ of 0.601. 
% Compared with + Retrieval, + Retrieval with the \textsc{Verifier} increases micro F$_1$ from 0.672 to 0.674 and macro F$_1$ from 0.595 to 0.601, although neither difference is statistically significant. Adding the \textsc{Verifier} increases micro F$_1$ from 0.662 to 0.676 and macro F$_1$ from 0.579 to 0.597 ($p<0.01$ for both comparisons). 
Appendix Table~\ref{tab:label_breakdown_semrep} provides a per-label performance breakdown on SemRepGS for the best micro-F$_1$ configuration (+ KB + Retrieval + \textsc{Verifier}).

\paragraph{\textbf{DDI}} 
A similar pattern is observed on the DDI dataset (Table~\ref{tab:test_results_ddi_chemprot}). KB context alone yields only marginal gains over the \textsc{Baseline}, increasing micro-F$_1$ from 0.769 to 0.774 with little change in macro-F$_1$ (0.595 $\rightarrow$ 0.596). Demonstration retrieval (+ Retrieval) significantly improves micro-F$_1$ to 0.841 and macro-F$_1$ to 0.670 ($p<0.01$ versus the \textsc{Baseline} for both metrics). Applying the \textsc{Verifier} further increases micro-F$_1$ to 0.881 and macro-F$_1$ to 0.706 ($p<0.01$ versus + Retrieval for both metrics), yielding the best performance among the evaluated configurations. 

KB context alone does not produce a statistically significant improvement over the \textsc{Baseline}. Combining demonstration retrieval with KB context also does not improve upon + Retrieval, yielding slightly lower micro-F$_1$ (0.839 versus 0.841) and macro-F$_1$ (0.661 versus 0.670). Adding the \textsc{Verifier} significantly increases micro-F$_1$ from 0.839 to 0.872 and macro-F$_1$ from 0.661 to 0.688 ($p<0.01$ for both metrics). However, these scores remain lower than those of + Retrieval + \textsc{Verifier} without KB context. Appendix Table~\ref{tab:label_breakdown_ddi} provides a per-label performance breakdown on DDI dataset for the + Retrieval + \textsc{Verifier} configuration.

% \textcolor{red}{Can you clarify why results are reported differently in Tables 2 and 3? I had commented on this before.}

\begin{table*}[t]
\centering
\small
\renewcommand{\arraystretch}{1.20}
\makebox[\linewidth][c]{%
\resizebox{0.90\linewidth}{!}{%

\begin{tabular}{lcccc}
\toprule
\multirow{2}{*}{Configuration} &
\multirow{2}{*}{Micro-F$_1$} &
\multicolumn{3}{c}{Macro} \\
\cmidrule(lr){3-5}
& &
Precision &
Recall &
F$_1$ \\
\midrule

\multicolumn{5}{l}{\textit{\textbf{DDI}}} \\
\midrule

\textsc{Baseline}
& \makecell{0.769 \\ {[0.758, 0.781]}}
& \makecell{0.528 \\ {[0.505, 0.552]}}
& \makecell{0.799 \\ {[0.779, 0.822]}}
& \makecell{0.595 \\ {[0.572, 0.618]}} \\

+ Retrieval
& \makecell{0.841$^{***}$ \\ {[0.831, 0.851]}}
& \makecell{0.609$^{***}$ \\ {[0.582, 0.637]}}
& \makecell{\textbf{0.814}$^{***}$ \\ {[0.793, 0.837]}}
& \makecell{0.670$^{***}$ \\ {[0.646, 0.695]}} \\

+ Retrieval w/\textsc{Verifier}
& \makecell{\textbf{0.881}$^{***\dagger\dagger\dagger}$ \\
  {[0.873, 0.890]}}
& \makecell{\textbf{0.668}$^{***\dagger\dagger\dagger}$ \\
  {[0.638, 0.696]}}
& \makecell{0.791$^{*\dagger\dagger\dagger}$ \\
  {[0.767, 0.815]}}
& \makecell{\textbf{0.706}$^{***\dagger\dagger\dagger}$ \\
  {[0.681, 0.732]}} \\

\addlinespace[0.6em]

+ KB
& \makecell{0.774 \\ {[0.763, 0.785]}}
& \makecell{0.526 \\ {[0.503, 0.549]}}
& \makecell{0.798 \\ {[0.777, 0.820]}}
& \makecell{0.596 \\ {[0.572, 0.618]}} \\

+ KB + Retrieval
& \makecell{0.839$^{***}$ \\ {[0.830, 0.849]}}
& \makecell{0.590$^{***}$ \\ {[0.565, 0.616]}}
& \makecell{\textbf{0.812}$^{***}$ \\ {[0.790, 0.835]}}
& \makecell{0.661$^{***}$ \\ {[0.637, 0.684]}} \\

+ KB + Retrieval w/\textsc{Verifier}
& \makecell{0.872$^{***\dagger\dagger\dagger}$ \\
  {[0.863, 0.881]}}
& \makecell{0.640$^{***\dagger\dagger\dagger}$ \\
  {[0.612, 0.669]}}
& \makecell{0.773$^{***\dagger\dagger\dagger}$ \\
  {[0.747, 0.797]}}
& \makecell{0.688$^{***\dagger\dagger\dagger}$ \\
  {[0.661, 0.713]}} \\

\midrule
\multicolumn{5}{l}{\textit{\textbf{ChemProt}}} \\
\midrule

\textsc{Baseline}
& \makecell{0.937 \\ {[0.929, 0.946]}}
& \makecell{0.912 \\ {[0.899, 0.925]}}
& \makecell{\textbf{0.939} \\ {[0.930, 0.948]}}
& \makecell{0.924 \\ {[0.913, 0.934]}} \\

+ Retrieval
& \makecell{0.940 \\ {[0.931, 0.949]}}
& \makecell{0.915 \\ {[0.900, 0.930]}}
& \makecell{\textbf{0.939} \\ {[0.927, 0.949]}}
& \makecell{0.924 \\ {[0.911, 0.936]}} \\

+ KB
& \makecell{0.939 \\ {[0.931, 0.947]}}
& \makecell{\textbf{0.916} \\ {[0.902, 0.928]}}
& \makecell{\textbf{0.939} \\ {[0.928, 0.948]}}
& \makecell{\textbf{0.926} \\ {[0.914, 0.936]}} \\

+ KB + Retrieval
& \makecell{\textbf{0.941} \\ {[0.933, 0.949]}}
& \makecell{0.913 \\ {[0.899, 0.926]}}
& \makecell{\textbf{0.939} \\ {[0.929, 0.949]}}
& \makecell{0.924 \\ {[0.913, 0.935]}} \\

\bottomrule
\end{tabular}%
}}
% \vspace{1mm}
% \begin{minipage}{0.90\linewidth}
% \scriptsize
% Numbers in brackets denote 95\% confidence intervals from bootstrap evaluation ($n=1{,}000$). Statistical significance is calculated using a paired bootstrap tests against the \textsc{Baseline} ($^{*}p<0.1$, $^{***}p<0.01$) and, for configurations with the \textsc{Verifier}, against the corresponding configuration without the \textsc{Verifier} ($^{\dagger\dagger\dagger}p<0.01$).
% \end{minipage}
\caption{Evaluation results on the DDI and ChemProt test sets. Because the evaluation includes all labels in these single-label multiclass benchmarks, micro-precision, micro-recall, and micro-F$_1$ are identical; therefore, only micro-F$_1$ is reported. Numbers in brackets denote 95\% confidence intervals from bootstrap evaluation ($n=1{,}000$). Statistical significance is calculated using a paired bootstrap tests against the \textsc{Baseline} ($^{*}p<0.1$, $^{***}p<0.01$) and, for configurations with the \textsc{Verifier}, against the corresponding configuration without the \textsc{Verifier} ($^{\dagger\dagger\dagger}p<0.01$).}
\label{tab:test_results_ddi_chemprot}
\end{table*}

\paragraph{\textbf{ChemProt}} 
Performance differences on ChemProt are more modest (Table~\ref{tab:test_results_ddi_chemprot}). Incorporating KB context alone produces only minor changes relative to the \textsc{Baseline}, increasing micro-F$_1$ from 0.937 to 0.939 and macro-F$_1$ from 0.924 to 0.926. Demonstration retrieval without KB context increases micro-F$_1$ from 0.937 to 0.940, while macro-F$_1$ remains unchanged at 0.924. Combining retrieval with KB context produces the highest micro-F$_1$ of 0.941, and macro-F$_1$ of 0.924, slightly below the 0.926 obtained with KB context alone. The observed differences from the \textsc{Baseline} are small. Because ChemProt evaluation focuses on discrimination among positive relation labels, retrieved demonstrations contain only supporting (positive) examples rather than contrastive \textsc{norel} cases. 
% \textcolor{red}{Is it KG grounding or KB grounding? Be consistent how you refer to it throughout (probably KB). This applies to the tables as well. Why KG there?} 
Appendix Table~\ref{tab:label_breakdown_chemprot} provides a per-label performance breakdown for the + KB + Retrieval configuration.

\subsection{Post-cutoff Temporal Precision Evaluation}
Using the best-performing configuration, \textsc{ANCHOR-RE} predicted 824 of the 3,890 candidate instances as containing a positive relation. Expert evaluation of 500 randomly sampled positive predictions showed that 69\% were correct. This precision exceeds the SemRepGS precision reported above (micro-precision = 0.576; macro-precision = 0.512). The higher precision may partially reflect differences in entity annotation quality: the temporal evaluation used PubTator3 rather than the MetaMap-based entity recognition used for SemRepGS, trading entity coverage for entity recognition accuracy. These results suggest that the framework not only maintains precision on newly published biomedical literature but also can accommodate outputs from different biomedical entity recognition systems.

\subsection{Generalization Across Open-weight LLM Backbones}
To examine generalization beyond the proprietary LLM backbone, we evaluate \textsc{ANCHOR-RE} using Qwen3 (Table~\ref{tab:qwen_all}). Qwen3 yields lower absolute scores than GPT-5 on all three benchmarks, both under direct prompting and with the best-performing configurations. Compared with direct prompting, the best-performing \textsc{ANCHOR-RE} configuration increases micro-F$_1$ from 0.571 to 0.577 on SemRepGS, from 0.567 to 0.772 on DDI, and from 0.914 to 0.925 on ChemProt. 

The contribution of each component varies across datasets. On DDI, Qwen3 achieves the best results with + KB + Retrieval + \textsc{Verifier}. On SemRepGS, retrieval alone provides little improvement, while KB context alone decreases micro-F$_1$ and slightly increases macro-F$_1$. On ChemProt, retrieval alone achieves the highest scores, and adding KB context does not provide an additional improvement.

We also examined whether pairing different LLM backbones as \textsc{decider} and \textsc{verifier} agents confers any benefits. The results are reported in~\ref{app:cross-family} and show that the prediction and verification stages can use different model families, but model capacity rather than model combination drives the performance: among the evaluated combinations, using GPT-5 for both stages yields the highest scores.

\subsection{Comparison with Prior Work}
Table \ref{tab:comparison} compares \textsc{ANCHOR-RE} with representative prior methods based on their reported results. To the best of our knowledge, \textsc{ANCHOR-RE} outperforms previously reported inference-only approaches while achieving performance competitive with supervised and instruction-tuned methods. 

\begin{table*}[t]
\centering
\small
\renewcommand{\arraystretch}{1.20}
\resizebox{\linewidth}{!}{%
\begin{tabular}{@{}lllcc@{}}
\toprule
\textbf{Dataset} & \textbf{Method} & \textbf{Strategy} & \textbf{Micro-F$_1$} & \textbf{Macro-F$_1$} \\
\midrule

\multirow{2}{*}{SemRepGS}
& Ming et al.~\citep{ming2024enhancing} & Fine-tuning & \textbf{0.702}  & 0.620 \\
& \textbf{ANCHOR-RE} & Inference only & 0.676 & 0.597 \\
\midrule

\multirow{4}{*}{DDI}
& Yuan et al.~\citep{yuan2024biomedical} & Prompt tuning + fine-tuning & 0.838 & 0.768 \\
& Jia et al.~\citep{jia2025meta} & Fine-tuning & 0.878 & -- \\
& Nachid-Idrissi et al.~\citep{nachid2025exploring} & Inference only & 0.693 & -- \\
& \textbf{ANCHOR-RE} & Inference only & \textbf{0.881} & 0.706 \\
\midrule

\multirow{6}{*}{ChemProt}
% & Yuan et al.~\citep{yuan2024biomedical} & Prompt tuning + fine-tuning & 0.805 & 0.771 \\
& Nachid-Idrissi et al.~\citep{nachid2025exploring} & Fine-tuning & 0.934 & -- \\
& Nachid-Idrissi et al.~\citep{nachid2025exploring} & Inference only & 0.766 & -- \\
& Zhang et al.~\citep{zhang2024study} & Inference only & 0.820 & -- \\
& Zhou et al.~\citep{zhou2024leap} & Instruction tuning & \textbf{0.951} & -- \\
& \textbf{ANCHOR-RE} & Inference only & 0.941 & 0.924 \\
\bottomrule
\end{tabular}%
}
\caption{
Comparison of \textsc{ANCHOR-RE} with representative supervised, instruction-tuned, and inference-only methods on the SemRepGS, DDI, and ChemProt benchmarks. For SemRepGS, micro- and macro-F$_1$ are computed over the positive relation classes only. Macro-F$_1$ values are omitted when unavailable in the original publications.
}
\label{tab:comparison}
\end{table*}

\section{Discussion}
Across three benchmarks, \textsc{ANCHOR-RE} demonstrates that structured knowledge and explicit verification can improve the reliability of LLM-based BioRE without task-specific fine-tuning or parameter updates. The performance gains are driven primarily by reductions in false-positive predictions, whereas the contribution of structured knowledge varies with the availability of relevant relational evidence. Evaluation on Qwen3 suggests that the framework's benefits are not limited to the proprietary GPT backbone, although absolute performance depends on the underlying LLM backbone. The remaining errors highlight challenges involving implicitly expressed and ambiguous relations. 

\subsection{Controlling Over-Prediction through Verification} 
In BioRE, LLMs offer flexible contextual reasoning but may generate plausible yet unsupported relation predictions. This behavior reflects broader challenges associated with hallucination and overconfidence in generative LLMs \citep{10.1145/3703155}. \textsc{ANCHOR-RE} mitigates this limitation by integrating symbolic knowledge (ontological constraints and structured KB evidence) into LLM reasoning, followed by a data-driven verification stage that verifies predictions using learned soft constraints. This design is particularly effective for \textsc{norel}-dominant datasets such as DDI and SemRepGS, where negative instances constitute 83\% and 68\% of candidate pairs, respectively. In these settings, reducing false-positive predictions leads to substantial performance improvements, consistently outperforming the direct prompting baseline.

A relation-level analysis further shows that the \textsc{Verifier} reduces false positive predictions across nearly all SemRepGS relation types. The largest reductions in false positive rate (FPR = FP / total negatives) are observed for relations with weak lexical cues, such as \textit{part\_of} (0.031$\rightarrow$0.019), \textit{coexists\_with} (0.033$\rightarrow$0.025), and \textit{location\_of} (0.036$\rightarrow$0.030). Unlike manually engineered rules, these verification rules are learned automatically from empirical error patterns observed during zero-shot LLM inference, allowing the framework to adapt to different relation schemas without manual rule design.
Representative examples are shown in Table~\ref{tab:error_patterns}. Reducing false positive predictions is particularly important for downstream applications such as biomedical knowledge graph construction and literature-based discovery, where erroneous relations may propagate and affect subsequent analyses.

\begin{table*}[h]
\centering
\scriptsize
\begin{tabular}{
  >{\raggedright\arraybackslash}p{2.1cm}
  >{\raggedright\arraybackslash}p{3.5cm}
  >{\raggedright\arraybackslash}p{3.2cm}
  >{\raggedright\arraybackslash}p{1.8cm}
}
\toprule
\textbf{Error Pattern} & \textbf{Verification Rejection Criterion} & \textbf{Example Sentence (Gold: \textsc{norel})} & \textbf{Incorrect  \textsc{Decider} Prediction} \\
\midrule

\textbf{1. Surface trigger ambiguity} 
& Causal lexical triggers align superficially with relation definitions, but the causal effect is mediated by another entity or complex. 
& Mut PML interacts with PML-\textbf{RARalpha}$_{subj}$ and potentiates PML-RARalpha-mediated inhibition of RA-dependent \textbf{transcription}$_{obj}$. 
& 
% \textit{RARalpha $\rightarrow$ transcription} predicted as 
\textit{disrupts} 
% & The causal effect is mediated by the PML–RARalpha complex rather than RARalpha alone. 
\\

\addlinespace

\textbf{2. Contextual co-occurrence} 
& Entities co-occur in sentences containing relation-like verbs, but no intrinsic semantic relation holds. 
& Detection of circulating \textbf{DNA}$_{subj}$ has been investigated for identifying various forms of \textbf{cancer}$_{obj}$. 
& 
% \textit{DNA $\rightarrow$ cancer} predicted as 
\textit{associated\_with} 
% & The sentence describes a diagnostic application rather than a biological association between DNA and cancer. 
\\

\addlinespace

\textbf{3. Directionality / polarity confusion} 
& Entities appear in parallel causal structures or outcome descriptions, leading to over-prediction of direct relations. 
& \textbf{Olanzapine}$_{subj}$ was associated with more discontinuation due to \textbf{weight gain}$_{obj}$ or metabolic effects. 
& 
% \textit{Olanzapine $\rightarrow$ weight gain} predicted as 
\textit{causes} 
% & The statement reports treatment discontinuation context rather than asserting a direct causal mechanism. 
\\

\bottomrule
\end{tabular}
\caption{Qualitative analysis of representative false-positive error patterns and how the \textsc{Verifier} corrects them. Error patterns denote high-level categories of failure, while each verification rejection criterion corresponds to a specific rule learned by the \textsc{Error Pattern Learner} and applied during verification.}
\label{tab:error_patterns}
\end{table*}

\subsection{Coverage of External KB Context} 
The effectiveness of external KB context depends on whether the underlying KB contains explicit relational evidence for the target entity pair. For SemRepGS, the KB context is constructed primarily from UMLS Metathesaurus concept linking and concept-level relations. Direct pair-level evidence is sparse, with only 7.16\% of instances containing explicit relations between the subject and object (Table \ref{tab:semrep_direct_kb}). On this subset with direct pair-level evidence, adding KB context increased micro-F$_1$ from 0.705 to 0.713 and macro-F$_1$ from 0.550 to 0.607 relative to the \textsc{Baseline}. Among instances without a direct relation, micro-F$_1$ remained unchanged, while macro-F$_1$ decreased slightly from 0.570 to 0.565. For DDI, DrugBank provides entity descriptions and information about interactions involving the target drugs and other drugs. Such neighboring interaction evidence is available for 57.4\% of DDI test instances. These findings suggest that external KB context provides greater benefits when relevant relational evidence is available, but offers limited benefits when such coverage is sparse.

Beyond coverage, KB evidence about an entity pair may be irrelevant to, or conflict with, the relation expressed in a particular sentence. Such concerns are related to knowledge conflicts in retrieval-augmented LLMs~\citep{jin2024tug}. We therefore examine sensitivity to unrelated KB context and corrupted relation labels in the + KB configuration. The perturbation procedures and complete results are provided in ~\ref{app:kb-robustness}. The results indicate that irrelevant KB evidence may be easier for the model to ignore than misleading evidence about the target evidence, and motivates future work for more selective KB context retrieval. 

\subsection{Generalizability of \textsc{ANCHOR-RE}} 
\textsc{ANCHOR-RE} improves upon direct prompting without task-specific fine-tuning, with gains observed across both proprietary and open-weight LLM backbones. The post-cutoff temporal evaluation further demonstrates precision on recently published biomedical literature, supporting its potential use in real-world biomedical literature mining. In comparison with representative BioRE methods reported in the literature, \textsc{ANCHOR-RE} outperforms the inference-only approach, although supervised and instruction-tuned models remain the strongest performers overall.

\begin{figure*}[h]
    \centering
    \includegraphics[width=\textwidth]{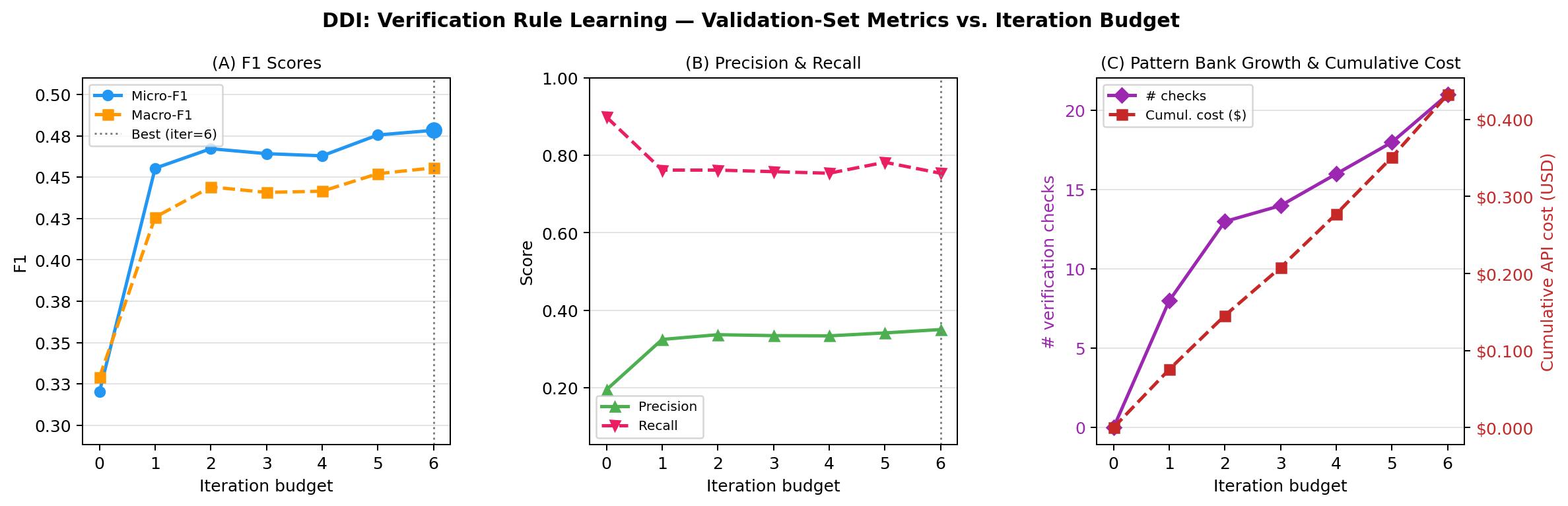}
    \vspace{-5pt}
    \caption{
    \small
    Performance and computational cost of offline verification rule learning across refinement iterations.
    }
    \label{fig:ddi}
    \vspace{-10pt}
\end{figure*}

\subsection{Practicality of Offline Error Pattern Learning}
Although \textsc{ANCHOR-RE} does not require task-specific fine-tuning, it introduces an offline error pattern learning stage to construct the \textit{Error Pattern Knowledge Bank}. To evaluate the efficiency of this stage and understand how the accumulated verification rules affect downstream performance, we analyze the relationship between refinement iterations, predictive performance, and construction cost. We analyze this process on the DDI benchmark, which provides the largest training corpus among the evaluated datasets. Following the experimental setup described in Section \ref{constructionmethod}, the original training set is divided into error pattern learning and validation splits to track performance throughout the refinement process.

Figure~\ref{fig:ddi} shows that both micro-F$_1$ and macro-F$_1$ improve rapidly during the first few refinement iterations before gradually converging, indicating that most useful verification knowledge is acquired early in the learning process. Meanwhile, the computational cost increases approximately linearly with the number of refinement iterations, false-positive cases, and generated verification rules, while remaining modest overall. Importantly, this cost is incurred only once for each dataset and backbone model. Once the \textit{Error Pattern Knowledge Bank} is constructed, the learned verification rules can be reused for subsequent inference without additional refinement, indicating that offline error pattern learning a lightweight preprocessing step rather than a recurring inference cost.

\subsection{Error Analysis on SemRepGS} 
Analysis of the remaining errors shows that most confusions occur between relations such as \textit{coexists\_with}, \textit{location\_of}, and \textit{process\_of} and the \textsc{norel} class. These relations are often expressed without explicit lexical triggers or involve ambiguous discourse cues, making them difficult to distinguish based on surface-level patterns alone. As summarized in Table~\ref{error}, the remaining errors fall into four main categories: (i) discourse-level relation misinterpretation, (ii) semantic or ontological relation confusion, (iii) implicit or multi-step biological reasoning failures, and (iv) long-range compositional reasoning failures. These findings suggest that, although the \textsc{Verifier} effectively mitigates over-prediction, further improvements will require stronger contextual grounding and enhanced compositional reasoning to better capture complex biological relationships.

\begin{table}[h]
\scriptsize
\centering
\resizebox{\textwidth}{!}{
\begin{tabular}{p{0.25\linewidth} p{0.55\linewidth} c}
\toprule
\textbf{Error Category} & \textbf{Representative Example (Subject \& Object with Semantic Types)} \\ 
\midrule
\textbf{Type 1: Discourse-level Relation Misinterpretation} & 
`Once \textbf{toxicity}$_{obj}$ [Injury or Poisoning] develops, supportive \textbf{therapy}$_{subj}$ [Therapeutic or Preventive Procedure] should be given, but there is no specific therapy to reduce toxicity or to enhance elimination.' \newline \textit{(Gold: none $\rightarrow$ Pred: treats)} 
\\
\midrule
\textbf{Type 2: Semantic or Ontological Relation Confusion} & 
`Clinically approved multi-\textbf{kinase inhibitors}$_{subj}$ [Amino Acid, Peptide, or Protein], such as nilotinib, inhibit DDR1-mediated \textbf{tumor growth}$_{obj}$ [Neoplastic Process] in xenograft models...' \newline \textit{(Gold: disrupts $\rightarrow$ Pred: prevents)} \\
\midrule
\textbf{Type 3: Implicit or Multi-step Biological Reasoning Failures} & 
`It was also suggested that spinosad-induced \textbf{free-radicals}$_{subj}$ [Biologically Active Substance] were eliminated by GSH-related antioxidant system in the \textbf{brain}$_{obj}$ [Body Part, Organ, or Organ Component] of Oreochromis niloticus.' \newline \textit{(Gold: part\_of $\rightarrow$ Pred: none)} \\
\midrule
\textbf{Type 4: Long-range Compositional Reasoning Failures} & 
`We recently showed that the \textbf{atypical PKC}$_{subj}$ [Amino Acid, Peptide, or Protein] (aPKC) isozyme PKCiota is overexpressed in human non-small cell lung cancer (NSCLC) cells and that PKCiota plays a critical role in the transformed growth of the human lung adenocarcinoma A549 cell line in vitro and \textbf{tumorigenicity}$_{obj}$ [Pathologic Function] in vivo.' \newline \textit{(Gold: none $\rightarrow$ Pred: causes)} \\
\bottomrule
\end{tabular}
}
\caption{Error type classification. Target entities are highlighted in \textbf{bold} with subscripts indicating their roles (\textbf{MENTION}$_{subj}$ for subject, \textbf{MENTION}$_{obj}$ for object), followed by their corresponding UMLS semantic types in brackets.} 
\label{error}
\end{table}

\subsection{Limitations and Future Work}
The current verification framework is designed as a one-sided rejection mechanism that primarily reduces false positive predictions. Future work could extend the \textsc{Verifier} beyond false-positive rejection to improve discrimination among semantically similar positive relation types, for example \textsc{treats} versus \textsc{prevents} or \textsc{coexists\_with} versus \textsc{associated\_with}. Second, our findings show that the effectiveness of KB grounding depends on the coverage of the underlying biomedical knowledge base and the quality of the retrieved evidence. Future work should investigate more selective evidence retrieval strategies that prioritize informative KB evidence while filtering irrelevant or potentially distracting context. Finally, the temporal evaluation assesses only predicted positives on their precision. Future work on recall estimation could provide a more comprehensive assessment of extraction performance.

Although \textsc{ANCHOR-RE} eliminates the need for task-specific fine-tuning and provides improved interpretability through explicit verification, it remains computationally more expensive than supervised transformer models once those models have been trained. Rather than viewing them as competing paradigms, future work may investigate hybrid triaging systems that allocate computational resources according to prediction confidence and downstream importance. For example, only predictions with low model confidence or relation types that are important for downstream applications (e.g., causal relations for biomedical knowledge graph construction) could be routed to \textsc{ANCHOR-RE} for knowledge-grounded reasoning and verification, while high-confidence predictions would be handled by lightweight supervised models.

\section{Conclusion}
BioRE remains challenging in realistic settings due to the scarcity of annotated data and the predominance of \textsc{norel} candidate pairs. In this work, we presented \textsc{ANCHOR-RE}, a training-free, agentic framework that improves the reliability of LLM-based BioRE through KB grounding, demonstration retrieval, and explicit verification. Results across three BioRE benchmarks show that the contribution of retrieval, KB context, and verification varies across datasets and configurations. Furthermore, the proposed framework generalizes across both proprietary and open-weight LLMs and remains effective under a post-cutoff temporal evaluation. Future work could investigate hybrid triaging strategies that combine the computational efficiency of supervised encoder models with the reasoning capabilities of \textsc{ANCHOR-RE}.

\section*{Acknowledgments}
This work was supported by funding from the National Center for Complementary and Integrative Health
(NCCIH) and the Office of Data Science Strategy (ODSS) (grant number: U01AT012871). Its contents
are solely the responsibility of the authors and do not necessarily represent the official views of the
NCCIH, ODSS, and the NIH.

\section*{Author Contributions}
\label{author_contributions}
SM: Conceptualization, Methodology, Software, Visualization, Validation, Formal Analysis, Writing – original draft, and Writing – review and editing; YH: Methodology, Software, Visualization, Validation, Formal Analysis, Writing – original draft, and Writing – review and editing; GH: Methodology, Software, Validation, Writing – original draft, and Writing – review and editing; RZ: Funding acquisition, Writing – review and editing; HK: Conceptualization, Methodology, Data Curation, Funding acquisition, Supervision, Writing – original draft, and Writing – review and editing.

\bibliographystyle{elsarticle-num-names} 
\bibliography{cas-refs}

%% else use the following coding to input the bibitems directly in the
%% TeX file.

% \begin{thebibliography}{00}

% %% \bibitem[Author(year)]{label}
% %% Text of bibliographic item

% \bibitem[ ()]{}
\appendix
\setcounter{table}{0}

\section{Supplementary Methods}
\subsection{Prompts Used in \textsc{ANCHOR-RE}}
\label{app:prompts}

% \textcolor{red}{I'd prefer using boxes with headers like you did in COLM paper, rather than using subsections for all these prompts.}
This section presents the complete prompts used for the direct-prompting
\textsc{Baseline}, \textsc{Decider}, \textsc{Verifier}, and \textsc{Error Pattern Learner} agents. The baseline and \textsc{Decider} agent prompts share the same instruction, except the shaded block below that highlights the contextual inputs added to the \textsc{Decider} prompt relative to the \textsc{Baseline}.

\begin{promptbox}{Prompt for \textsc{Baseline} (and \textsc{Decider}) agents}
\label{app:baseline_prompt}
%\label{deciderprompt}

You are an expert Biomedical Information Extraction Agent. Your primary
function is to accurately extract semantic relationships between biomedical entities from scientific text.

\promptheading{Task Description}
Given a biomedical sentence and a target pair of entities (SUBJECT and OBJECT), determine which biomedical relation label best describes their relationship. Use the provided sentence together with the relation definitions to determine the most appropriate relation. If the sentence expresses no explicit or verifiable relationship, output \texttt{none}.

\promptheading{Linguistic and Semantic Guidance}

\begin{enumerate}

\item \textbf{Trigger Identification}

\begin{itemize}
\item Verbs indicate causal or action-based relations (e.g., \textit{inhibits}, \textit{causes}, \textit{prevents}, \textit{treats}).
\item Nominalizations may express events (e.g., \textit{treatment}, \textit{activation}, \textit{expression}).
\item Prepositions often indicate static associations or locations.
\item Implicit triggers may occur in noun phrases (e.g., ``aspirin therapy'' implies \textsc{TREATS}).
\end{itemize}

\item \textbf{Trigger Positioning}

\begin{itemize}
\item Verbs typically occur between SUBJECT and OBJECT.
\item Nominalizations may occur before, between, or after the entity pair.
\item Prepositions frequently precede the OBJECT.
\item Carefully resolve coordinated entities connected by conjunctions.
\end{itemize}

\item \textbf{Entity-Type Heuristics}

\begin{itemize}
\item Pharmacologic Substance $\rightarrow$ Disease:
      \textsc{TREATS}, \textsc{PREVENTS}, \textsc{CAUSES}
\item Gene / Protein $\rightarrow$ Chemical / Protein:
      \textsc{INTERACTS\_WITH},
      \textsc{ASSOCIATED\_WITH},
      \textsc{STIMULATES},
      \textsc{INHIBITS}
\item Disease $\rightarrow$ Finding:
      \textsc{ASSOCIATED\_WITH},
      \textsc{CAUSES},
      \textsc{COEXISTS\_WITH}
\item Anatomical Structure $\rightarrow$ Disease:
      \textsc{LOCATION\_OF},
      \textsc{PART\_OF},
      \textsc{PROCESS\_OF}
\end{itemize}

\item \textbf{Decision Rules}

\begin{itemize}
\item Identify the trigger phrase connecting SUBJECT and OBJECT.
\item Consider both linguistic context and semantic types.
\item Predict the most specific relation supported by the evidence.
\item Output \texttt{none} if
      \begin{itemize}
      \item no explicit relation is expressed,
      \item the statement is purely descriptive,
      \item the connection is semantically unsupported
      \end{itemize}
\end{itemize}

\end{enumerate}

\promptheading{Required JSON Output}

\begin{lstlisting}[
  language=json,
  basicstyle=\ttfamily\scriptsize,
  breaklines=true,
  columns=fullflexible
]
{
  "subject": "<exact entity text>",
  "subject_type": "<entity type>",
  "object": "<exact entity text>",
  "object_type": "<entity type>",
  "trigger_span": "<supporting phrase>",
  "frame": "<brief explanation>",
  "label": "<candidate relation or 'none'>"
}
\end{lstlisting}

\vspace{0.5em}

\textbf{Input}

\begin{itemize}
    \item \textbf{Subject:} [SUBJECT ENTITY MENTION] (type: [SUBJECT ENTITY TYPE])
    \item \textbf{Object:} [OBJECT ENTITY MENTION] (type: [OBJECT ENTITY TYPE])
    \item \textbf{Sentence:} [SENTENCE WHERE SUBJECT AND OBJECT MENTIONS ARE LOCATED]
    \item \textbf{Candidate Relations:} [LIST OF CANDIDATE RELATION TYPES AND THEIR DEFINITIONS]
    % \item \textbf{Retrieved Demonstrations:} [LIST OF RETRIEVED EXAMPLES]   Reason over the examples above, paying particular attention to the negative examples labeled \texttt{none}.
    % \item \textbf{Knowledge Base Evidence:} [EXTRACTED KNOWLEDGE FOR SUBJECT AND OBJECT]
        \begin{tcolorbox}[
      colback=gray!8,
      colframe=gray!45,
      boxrule=0.5pt,
      arc=1pt,
      left=4pt,
      right=4pt,
      top=3pt,
      bottom=3pt,
      title={Additional context provided to the \textsc{Decider}},
      fonttitle=\bfseries\footnotesize
    ]
    \begin{itemize}
        \item \textbf{Retrieved Demonstrations:}
        [LIST OF RETRIEVED EXAMPLES] Reason over the examples above,
        paying particular attention to the negative examples labeled
        \texttt{none}.
    
        \item \textbf{Knowledge Base Evidence:}
        [EXTRACTED KNOWLEDGE FOR SUBJECT AND OBJECT]
    \end{itemize}
    \end{tcolorbox}

\end{itemize}

\textbf{Decision Guidelines}

\begin{enumerate}
    \item Consider \texttt{none} as the default prediction.
    \item Assign a relation label only if the sentence contains a clear and explicit trigger connecting the subject and object.
    \item If the evidence is ambiguous, uncertain, or not directly supported by the sentence, predict \texttt{none}.
\end{enumerate}

\vspace{0.5em}

\textbf{Question}

What relation, if any, is expressed between the subject and object in the given sentence?
\end{promptbox}

\begin{promptbox}{\textsc{Verifier} agent prompt}
\label{Verifier}

You are a biomedical relation verification expert.

\textbf{Decision Rubric}

\textbf{1. Evidence-first}

\begin{itemize}
    \item If the sentence explicitly states or strongly implies the predicted relation between the subject and object, \textbf{ACCEPT} the prediction.
    \item Quote the strongest supporting phrase as evidence.
\end{itemize}

\textbf{2. Apply Verification Checks}

\begin{itemize}
    \item Only \textbf{REJECT} the prediction if \emph{both} conditions are satisfied:
    \begin{enumerate}
        \item no strong supporting evidence can be quoted from the sentence; and
        \item at least one verification check is clearly supported by the sentence.
    \end{enumerate}
    \item If the decision is uncertain or borderline, \textbf{ACCEPT} the prediction.
\end{itemize}

\textbf{Input}

\begin{itemize}
    \item \textbf{Sentence:} [SENTENCE WHERE SUBJECT AND OBJECT MENTIONS ARE LOCATED IN]
    \item \textbf{Subject:} [SUBJECT ENTITY MENTION] ([SUBJECT ENTITY TYPE])
    \item \textbf{Object:} [OBJECT ENTITY MENTION] ([OBJECT ENTITY TYPE])
    \item \textbf{Predicted Relation:} [PREDICTED RELATION TYPE]
    \item \textbf{Relation Definition:} [DEFINITION OF PREDICTED RELATION TYPE]
    \item \textbf{Verification Checks for \texttt{<PREDICTED RELATION TYPE>}:} [VERIFICATION RULES]

\end{itemize}
\promptheading{Required JSON Output}

\begin{lstlisting}[
  language=json,
  basicstyle=\ttfamily\scriptsize,
  breaklines=true,
  columns=fullflexible
]
{
  "evidence": "<quoted phrase from the sentence, or 'none found'>",
  "final_label": "<predicted relation if accepted, otherwise 'none'>",
  "reasoning": "<one-sentence explanation>"
}
\end{lstlisting}

\end{promptbox}

\begin{promptbox}{\textsc{Error Pattern Learner} agent prompt}
\label{learner}

The \textsc{Error Pattern Learner} uses an initialization prompt during the first iteration and a refinement prompt during subsequent iterations.

\promptheading{Initialization Prompt:}

You are creating verification checks for the label \texttt{<LABEL>} against false positives ("no relation").

Definition: \texttt{<DEFINITIONS>}

True Positives:
\begin{itemize}
\item TP1: Sentence, Subject, Object
\item TP2: ...
\end{itemize}

False Positives:
\begin{itemize}
\item FP1: Sentence, Subject, Object, Decider reasoning, Evidence span
\item FP2: ...
\end{itemize}

Your Tasks:
\begin{enumerate}
\item Describe 3--5 conceptual semantic properties that must hold for relation \texttt{<LABEL>}.
\item Group false positives into 2--4 conceptual error categories.
\item Generate at most three verification checks satisfying:
\begin{itemize}
\item Each check may cover multiple related error categories.
\item Store covered categories in \texttt{negative\_patterns}.
\item Questions should target semantic failures rather than lexical patterns.
\item No check should reject any true positive.
\end{itemize}
\end{enumerate}

\promptheading{Required JSON Output}

\begin{lstlisting}[
  language=json,
  basicstyle=\ttfamily\scriptsize,
  breaklines=true,
  columns=fullflexible
]
{
  "positive_patterns": ["<semantic property 1>", "<semantic property 2>", ...],
  "new_checks": [
    {
      "pattern_id": "<LABEL>_NON_CONFUSION_iter1_<N>",
      "negative_patterns": ["<FP subcategory this check targets>", 
      "<another subcategory if combined with OR>"],
      "question": "<broad YES/NO question covering all listed subcategories>",
      "if_yes": "flip_to_none",
      "if_no": "accept"
    }
  ],
  "check_revisions": [],
  "positive_pattern_revisions": []
}
\end{lstlisting}
% \end{promptbox}

\promptheading{Refinement Prompt:}

Existing Checks:
\begin{itemize}
\item Pattern 1: ...
\item Pattern 2: ...
\end{itemize}

New Batch of False Positives:
\begin{itemize}
\item FP1: Sentence, Subject, Object, Decider reasoning, Evidence span
\item FP2: ...
\end{itemize}

True Positives:
\begin{itemize}
\item TP1: Sentence, Subject, Object
\item TP2: ...
\end{itemize}

Your Tasks: 
\begin{enumerate}
    \item Revise existing positive patterns if new FPs reveal they are too broad or incomplete. Use "positive\_pattern\_revisions" with the index and revised text. Leave [] if unchanged.
    \item Broaden existing checks if a new FP represents the same semantic failure as an existing check's negative\_pattern but the current question would NOT catch it (too narrow). Use "check\_revisions" with the pattern\_id, a broader revised\_question, and optionally revised\_negative\_patterns if the category description should be widened too. Leave [] if no broadening needed.
    \item Add new checks for genuinely new FP categories not covered by any existing check. Each new check lists ALL the FP subcategories it covers in "negative\_patterns" (array). The question must be OVERARCHING — use OR to combine related subcategories when possible. Leave "new\_checks" as [] if everything is already covered. Each new check MUST NOT fire on any true positive.
\end{enumerate}

\promptheading{Required JSON Output}

\begin{lstlisting}[
  language=json,
  basicstyle=\ttfamily\scriptsize,
  breaklines=true,
  columns=fullflexible
]
{
  "positive_pattern_additions": ["<new positive property, if any>"],
  "positive_pattern_revisions": [{"index": <int>, "revised": "<revised text>"}],
  "check_revisions": [
    {
      "pattern_id": "<existing pattern_id>",
      "revised_question": "<broader YES/NO question>",
      "revised_negative_patterns": ["<updated FP subcategory list>"],
      "revision_note": "<what new FP this now catches>"
    }
  ],
  "new_checks": [
    {
      "pattern_id": "<LABEL>_NON_CONFUSION_iter<N>_<ID>",
      "negative_patterns": ["<FP subcategory 1>", "<FP subcategory 2 
      if combined with OR>"],
      "question": "<broad YES/NO question covering all listed subcategories; 
      YES => flip_to_none>",
      "if_yes": "flip_to_none",
      "if_no": "accept"
    }
  ]
}
\end{lstlisting}

\end{promptbox}

\subsection{KB Evidence Format Examples}
\label{app:kg-examples}

Figure~\ref{fig:kg-all} shows representative examples of the structured KB evidence blocks used across datasets, as formatted in the \textsc{Decider} prompt. Each block is dataset-specific and reflects the information retrieved from the corresponding external knowledge resource.

\begin{figure*}[!htbp]
\scriptsize
\centering

\textbf{(a) SemRepGS -- UMLS evidence block (subject: \textit{Prostate Cancer}, object: \textit{PSA})}

\begin{verbatim}
KB Report for Prostate Cancer -> PSA:
Direct Relations:
  - ASSOCIATED_WITH
Direct CUI evidence:
  - Prostate Cancer -> prostate specific antigen: ASSOCIATED_WITH
  - Prostate Cancer -> PSA: ASSOCIATED_WITH
Possible relations Prostate Cancer -> PSA:
  - PROCESS_OF, AFFECTS, ASSOCIATED_WITH
Related concepts of Prostate Cancer:
  - Prostate Neoplasms
  - Prostatic Intraepithelial Neoplasia
  - Benign Prostatic Hyperplasia
Related concepts of PSA:
  - Biomarkers, Tumor
  - kininogenase
  - KLK3 gene
\end{verbatim}

\vspace{1pt}

\textbf{(b) DDI -- DrugBank evidence block (subject: \textit{nalidixic acid}, object: \textit{quinidine})}

\begin{verbatim}
DrugBank KB report for nalidixic acid <-> quinidine:
Subject: Nalidixic acid (DrugBank ID: DB00779)
  Class: Diazanaphthalenes
  Indication: For the treatment of urinary tract infections caused by
    susceptible gram-negative microorganisms, including E. Coli,
    Enterobacter species, Klebsiella species, and Proteus species.
  Mechanism: The active metabolite hydroxynalidixic acid binds strongly,
    but reversibly, to DNA, interfering with RNA and protein synthesis.
  Related interactions:
    - Ivabradine: Ivabradine may increase the QTc-prolonging activities
      of Nalidixic acid.
    - Melphalan: The risk or severity of gastrointestinal bleeding can be
      increased when Nalidixic acid is combined with Melphalan.
    - Didanosine: Didanosine can cause a decrease in the absorption of
      Nalidixic acid resulting in a reduced serum concentration.
Object: Quinidine (DrugBank ID: DB00908)
  Class: Cinchona alkaloids
  Indication: Quinidine is indicated for the management and prophylactic
    therapy of atrial fibrillation/flutter and ventricular arrhythmias.
  Mechanism: Quinidine blocks the rapid sodium channel (INa) in Purkinje
    fibers, decreasing the phase zero of rapid depolarization. It also
    reduces repolarizing K+ currents and the L-type calcium current.
  Related interactions:
    - Afatinib: The serum concentration of Afatinib can be increased
      when it is combined with Quinidine.
    - Modafinil: The metabolism of Quinidine can be increased when
      combined with Modafinil.
    - Armodafinil: The metabolism of Quinidine can be increased when
      combined with Armodafinil.
\end{verbatim}

\vspace{1pt}

\textbf{(c) ChemProt -- CTD evidence block (protein: \textit{AR}, chemical: \textit{bicalutamide})}

\begin{verbatim}
CTD KB report for AR <-> bicalutamide:
Entity1 (standardized): AR
Entity2 (standardized): bicalutamide
Relevant one-hop CTD interactions:
  1. Gene 'AR' -> Chemical '6-fluorotestosterone': bicalutamide inhibits
     the reaction [6-fluorotestosterone binds to and results in increased
     activity of AR protein]
  2. Gene 'AR' -> Chemical '10074-G5': 10074-G5 results in decreased
     expression of AR protein
  3. Chemical 'bicalutamide' -> Gene 'ABCB11': bicalutamide results in
     decreased activity of ABCB11 protein
  4. Chemical 'bicalutamide' -> Gene 'ACP3': bicalutamide results in
     decreased expression of ACP3 mRNA
  5. Chemical 'bicalutamide' -> Gene 'ACSL1': bicalutamide results in
     increased expression of ACSL1 mRNA
\end{verbatim}

\caption{Representative structured evidence blocks used to augment the \textsc{Decider} prompt}
\label{fig:kg-all}
\end{figure*}

\subsection{Dataset Label Distribution}
\label{app:data-stats}
Table~\ref{tab:label_distribution} reports the label distribution across train, dev/validation, and test splits for SemRepGS, DDI, and ChemProt. DDI (SemEval-2013) has no official development split. ChemProt labels are reported at the CPR subgroup level used for evaluation (cpr:3--cpr:9).

\begin{table*}[h]
\centering
\scriptsize
\setlength{\tabcolsep}{6pt}
\renewcommand{\arraystretch}{1.05}
\begin{tabular}{llrrrr}
\toprule
\textbf{Dataset} & \textbf{Label} & \textbf{Train} & \textbf{Dev} & \textbf{Test} \\
\midrule
\multirow{23}{*}{SemRepGS}
 & administered\_to   & 188   & 14  & 21   \\
 & affects            & 118   & 8   & 79   \\
 & associated\_with   & 278   & 10  & 12   \\
 & augments           & 91    & 11  & 12   \\
 & causes             & 165   & 11  & 43   \\
 & coexists\_with     & 290   & 25  & 24   \\
 & compared\_with     & 74    & 11  & 8    \\
 & diagnoses          & 155   & 14  & 16   \\
 & disrupts           & 105   & 9   & 12   \\
 & inhibits           & 311   & 24  & 12   \\
 & interacts\_with    & 1,030 & 109 & 34   \\
 & isa                & 108   & 13  & 85   \\
 & location\_of       & 223   & 7   & 170  \\
 & part\_of           & 262   & 10  & 129  \\
 & precedes           & 98    & 12  & 13   \\
 & predisposes        & 88    & 17  & 11   \\
 & prevents           & 99    & 10  & 13   \\
 & process\_of        & 666   & 72  & 190  \\
 & produces           & 113   & 13  & 15   \\
 & stimulates         & 232   & 33  & 16   \\
 & treats             & 363   & 35  & 96   \\
 & uses               & 277   & 30  & 26   \\
 & \textsc{norel}               & 2,750 & 258 & 2,174 \\
\cmidrule{2-5}
 & \textbf{Total}     & \textbf{8,084} & \textbf{756} & \textbf{3,211} \\
\midrule
\multirow{6}{*}{DDI}
 & advise             & 826    & --- & 221   \\
 & effect             & 1,687  & --- & 360   \\
 & int                & 188    & --- & 96    \\
 & mechanism          & 1,319  & --- & 302   \\
 & \textsc{norel}               & 23,771 & --- & 4,737 \\
\cmidrule{2-5}
 & \textbf{Total}     & \textbf{27,792} & \textbf{---} & \textbf{5,716} \\
\midrule
\multirow{6}{*}{ChemProt}
 & cpr:3              & 777   & 552   & 667   \\
 & cpr:4              & 2,260 & 1,103 & 1,667 \\
 & cpr:5              & 170   & 116   & 198   \\
 & cpr:6              & 235   & 199   & 293   \\
 & cpr:9              & 727   & 457   & 644   \\
\cmidrule{2-5}
 & \textbf{Total}     & \textbf{4,169} & \textbf{2,427} & \textbf{3,469} \\
\bottomrule
\end{tabular}
\caption{Label distribution across dataset splits. ``---'' indicates no official split exists.}
\label{tab:label_distribution}
\end{table*}

\setcounter{table}{0}
\section{Supplementary Results}

\subsection{Cost Analysis}
\label{latencycost}

Table~\ref{tab:cost} summarizes the inference latency and computational cost under the best-performing configuration for each dataset. For GPT-5 (\texttt{gpt-5-mini-2025-08-07}), we report the API cost based on OpenAI pricing. For the open-weight Qwen3 (\texttt{Qwen3-32B}) model, we report the mean per-instance latency, including demonstration retrieval, relation prediction, and verification when invoked.

\begin{table}[h]
\centering
\scriptsize
\resizebox{\linewidth}{!}{%
\begin{tabular}{llcccc}
\toprule
\textbf{Backbone} &
\textbf{Dataset} &
\textbf{\#Instances} &
\textbf{Time / Inst (s)} &
\textbf{Cost / Inst (\$)} &
\textbf{Total Cost (\$)} \\
\midrule

\multirow{3}{*}{GPT-5}
& SemRepGS & 3,211 & 17.0 & 0.0017 & 5.71 \\
& DDI      & 5,716 & 12.7 & 0.0016 & 9.17 \\
& ChemProt & 3,469 & 7.8  & 0.0017 & 5.97 \\

\midrule

% \multirow{3}{*}{\texttt{Qwen3.5-2B}}
% & SemRepGS & 3,211 & 2.3 & -- & -- \\
% & DDI      & 5,716 & 1.9 & -- & -- \\
% & ChemProt & 3,469 & 1.4 & -- & -- \\

% \midrule

\multirow{3}{*}{Qwen3}
& SemRepGS & 3,211 & 5.4 & -- & -- \\
& DDI      & 5,716 & 4.1 & -- & -- \\
& ChemProt & 3,469 & 2.5 & -- & -- \\

\bottomrule
\end{tabular}%
}
\caption{Inference latency and cost under the best-performing ANCHOR-RE configuration. API costs are reported only for GPT-5.}
\label{tab:cost}
\end{table}

\subsection{Per-Label Performance Breakdown Across SemRepGS, DDI, and ChemProt}
\label{app:per}
Tables~\ref{tab:label_breakdown_semrep}, \ref{tab:label_breakdown_ddi}, and \ref{tab:label_breakdown_chemprot} present the per-label performance breakdown for SemRepGS, DDI, and ChemProt, respectively.

\begin{table}[h]
\centering
\small
\setlength{\tabcolsep}{6pt}
\renewcommand{\arraystretch}{1.1}
\begin{tabular}{lccc}
\toprule
\textbf{Label} & \textbf{Precision} & \textbf{Recall} & \textbf{F$_1$} \\
\midrule
administered\_to & 0.395 & 0.714 & 0.509 \\
affects & 0.510 & 0.658 & 0.575 \\
associated\_with & 0.117 & 0.583 & 0.194 \\
augments & 0.290 & 0.750 & 0.419 \\
causes & 0.629 & 0.907 & 0.743 \\
coexists\_with & 0.205 & 0.667 & 0.314 \\
compared\_with & 0.471 & 1.000 & 0.640 \\
diagnoses & 0.750 & 0.938 & 0.833 \\
disrupts & 0.545 & 0.500 & 0.522 \\
inhibits & 0.421 & 0.667 & 0.516 \\
interacts\_with & 0.346 & 0.794 & 0.482 \\
isa & 0.812 & 0.918 & 0.862 \\
location\_of & 0.660 & 0.817 & 0.730 \\
part\_of & 0.657 & 0.672 & 0.664 \\
precedes & 0.556 & 0.833 & 0.667 \\
predisposes & 0.400 & 0.364 & 0.381 \\
prevents & 0.867 & 1.000 & 0.929 \\
process\_of & 0.773 & 0.984 & 0.866 \\
produces & 0.316 & 0.429 & 0.364 \\
stimulates & 0.385 & 0.938 & 0.546 \\
treats & 0.667 & 0.842 & 0.744 \\
uses & 0.500 & 0.885 & 0.639 \\
\midrule
\textbf{Macro avg} & 0.512 & 0.766 & 0.597 \\
\textbf{Micro avg} & 0.576 & 0.816 & 0.676 \\
\bottomrule
\end{tabular}
\caption{Per-label performance on SemRepGS test set (positive labels only) under the best configuration (+ KB + Retrieval + \textsc{Verifier}).}
\label{tab:label_breakdown_semrep}
\end{table}

\begin{table}[h]
\centering
\small
\setlength{\tabcolsep}{6pt}
\renewcommand{\arraystretch}{1.1}
\begin{tabular}{lccc}
\toprule
\textbf{Label} & \textbf{Precision} & \textbf{Recall} & \textbf{F$_1$} \\
\midrule
advice & 0.624 & 0.955 & 0.755 \\
effect & 0.552 & 0.911 & 0.688 \\
int & 0.585 & 0.396 & 0.472 \\
mechanism & 0.593 & 0.801 & 0.682 \\
none & 0.984 & 0.891 & 0.935 \\
\midrule
\textbf{Macro avg} & 0.668 & 0.791 & 0.706 \\
\textbf{Weighted avg} & 0.912 & 0.881 & 0.889 \\
\bottomrule
\end{tabular}
\caption{Per-label performance on the DDI test set (including the \textsc{norel} class) under the best configuration (+ Retrieval + \textsc{Verifier}).}
\label{tab:label_breakdown_ddi}
\end{table}

\begin{table}[h]
\centering
\small
\setlength{\tabcolsep}{6pt}
\renewcommand{\arraystretch}{1.1}
\begin{tabular}{lccc}
\toprule
\textbf{Label} & \textbf{Precision} & \textbf{Recall} & \textbf{F$_1$} \\
\midrule
cpr:3 & 0.934 & 0.894 & 0.913 \\
cpr:4 & 0.962 & 0.956 & 0.959 \\
cpr:5 & 0.858 & 0.944 & 0.899 \\
cpr:6 & 0.829 & 0.959 & 0.889 \\
cpr:9 & 0.982 & 0.941 & 0.961 \\
\midrule
\textbf{Macro avg} & 0.913 & 0.939 & 0.924 \\
\textbf{Weighted avg} & 0.943 & 0.941 & 0.941 \\
\bottomrule
\end{tabular}
\caption{Per-label performance on the ChemProt test set under the best configuration (+ KB + Retrieval).}
\label{tab:label_breakdown_chemprot}
\end{table}

\subsection{Evaluation with Open-Weight LLM Backbones}
Detailed results for \textsc{ANCHOR-RE} with the Qwen3 backbones on SemRepGS, DDI, and ChemProt are presented below.

\begin{table*}[t]
\centering
\small
\setlength{\tabcolsep}{4pt}
\caption{Performance of \textsc{ANCHOR-RE} with Qwen3. SemRepGS excludes \textsc{norel} from micro- and macro-averaging. DDI includes all classes, including \textsc{norel}; ChemProt includes only the five retained positive classes. For DDI and ChemProt, micro-precision and micro-recall equal micro-F$_1$ and are omitted. Bold values indicate the highest score within each dataset. Confidence intervals and statistical significance tests are omitted for readability.}
\label{tab:qwen_all}
\begin{tabular}{lcccccc}
\toprule
& \multicolumn{3}{c}{Micro} & \multicolumn{3}{c}{Macro} \\
\cmidrule(lr){2-4}\cmidrule(lr){5-7}
Configuration & Precision & Recall & F$_1$ & Precision & Recall & F$_1$ \\
\midrule
\multicolumn{7}{l}{\textit{SemRepGS}} \\
\midrule
\textsc{Baseline} & 0.472 & 0.723 & 0.571 & 0.399 & 0.721 & 0.484 \\
+ Retrieval & 0.460 & \textbf{0.750} & 0.571 & 0.392 & \textbf{0.738} & 0.488 \\
+ Retrieval + \textsc{Verifier} & 0.473 & 0.729 & 0.574 & 0.399 & 0.717 & 0.489 \\
+ KB & 0.476 & 0.689 & 0.563 & 0.409 & 0.712 & 0.486 \\
+ KB + Retrieval & 0.467 & 0.746 & 0.575 & 0.414 & 0.719 & 0.503 \\
+ KB + Retrieval + \textsc{Verifier} & \textbf{0.477} & 0.729 & \textbf{0.577} & \textbf{0.423} & 0.706 & \textbf{0.506} \\
\midrule
\multicolumn{7}{l}{\textit{DDI}} \\
\midrule
\textsc{Baseline} & -- & -- & 0.567 & 0.398 & 0.735 & 0.435 \\
+ Retrieval & -- & -- & 0.698 & 0.462 & 0.766 & 0.528 \\
+ Retrieval + \textsc{Verifier} & -- & -- & 0.748 & 0.490 & \textbf{0.774} & 0.563 \\
+ KB & -- & -- & 0.593 & 0.404 & 0.732 & 0.448 \\
+ KB + Retrieval & -- & -- & 0.683 & 0.450 & 0.764 & 0.514 \\
+ KB + Retrieval + \textsc{Verifier} & -- & -- & \textbf{0.772} & \textbf{0.500} & 0.773 & \textbf{0.577} \\
\midrule
\multicolumn{7}{l}{\textit{ChemProt}} \\
\midrule
\textsc{Baseline} & -- & -- & 0.914 & 0.887 & 0.919 & 0.900 \\
+ Retrieval & -- & -- & \textbf{0.925} & \textbf{0.902} & \textbf{0.926} & \textbf{0.912} \\
+ KB & -- & -- & 0.912 & 0.884 & 0.919 & 0.898 \\
+ KB + Retrieval & -- & -- & 0.914 & 0.885 & 0.920 & 0.899 \\
\bottomrule
\end{tabular}
\end{table*}

\subsection{SemRepGS Performance by Direct KB Relation Availability}
\label{app:kb-coverage}
Table~\ref{tab:semrep_direct_kb} compares \textsc{Baseline} and + KB on SemRepGS, stratified by whether the retrieved KB context contains a direct relation between the target subject and object. 

\begin{table*}[t]
\centering
\small
\caption{SemRepGS performance stratified by whether the retrieved KB context contains a direct relation between the target subject and object. The \textsc{norel} class is excluded from micro-and macro-averaging.}
\label{tab:semrep_direct_kb}
\medskip
\setlength{\tabcolsep}{4pt}
\begin{tabular}{lcccccc}
\toprule
& \multicolumn{3}{c}{Micro} & \multicolumn{3}{c}{Macro} \\
\cmidrule(lr){2-4}\cmidrule(lr){5-7}
Configuration & Precision & Recall & F$_1$ & Precision & Recall & F$_1$ \\
\midrule

\multicolumn{7}{l}{
\textit{With a direct subject--object KB relation}} \\
\midrule
\textsc{Baseline} & 0.591 & 0.873 & 0.705 & 0.512 & 0.625 & 0.550 \\
+ KB & \textbf{0.596} & \textbf{0.889} & \textbf{0.713} & \textbf{0.560} & \textbf{0.716} & \textbf{0.607} \\

\midrule
\multicolumn{7}{l}{
\textit{Without a direct subject--object KB relation}} \\
\midrule
\textsc{Baseline} & \textbf{0.531} & 0.842 & \textbf{0.651} & \textbf{0.470} & \textbf{0.792} & \textbf{0.570} \\
+ KB & \textbf{0.531} & \textbf{0.843} & \textbf{0.651} & 0.466 & 0.788 & 0.565 \\

\bottomrule
\end{tabular}
\end{table*}

\subsection{Cross-Family Verification}
\label{app:cross-family}

We evaluate cross-family verification by pairing GPT-5 and Qwen3 as the \textsc{Decider} and \textsc{Verifier}. The learned verification patterns are kept fixed, and within each \textsc{Decider} group, the \textsc{Verifier} is applied to the same predictions from the + KB + Retrieval configuration.

\begin{table*}[t]
\centering
\small
\caption{Cross-family verification on SemRepGS. ``--'' denotes the + KB + Retrieval configuration without verification. Bold values indicate the highest score within each \textsc{Decider} group. The \textsc{norel} class is excluded from micro- and macro-averaging.}

\label{tab:cross_family_semrep}
\begin{tabular}{llcccccc}
\toprule
\textsc{Decider} & \textsc{Verifier}
& \multicolumn{3}{c}{Micro}
& \multicolumn{3}{c}{Macro} \\
\cmidrule(lr){3-5}\cmidrule(lr){6-8}
& & Precision & Recall & F$_1$
  & Precision & Recall & F$_1$ \\
\midrule
Qwen3 & --
& 0.467 & \textbf{0.746} & 0.575
& 0.414 & \textbf{0.719} & 0.503 \\
Qwen3 & Qwen3
& 0.477 & 0.729 & 0.577
& 0.423 & 0.706 & 0.506 \\
Qwen3 & GPT-5
& \textbf{0.523} & 0.716 & \textbf{0.605}
& \textbf{0.476} & 0.697 & \textbf{0.544} \\
\midrule
GPT-5 & --
& 0.541 & \textbf{0.853} & 0.662
& 0.480 & \textbf{0.791} & 0.579 \\
GPT-5 & Qwen3
& 0.572 & 0.743 & 0.646
& 0.501 & 0.728 & 0.573 \\
GPT-5 & GPT-5
& \textbf{0.576} & 0.816 & \textbf{0.676}
& \textbf{0.512} & 0.766 & \textbf{0.597} \\
\bottomrule
\end{tabular}
\end{table*}

\subsection{Sensitivity to Perturbed KB Context}
\label{app:kb-robustness}

\textbf{Irrelevant evidence.} We replaced each instance's retrieved KB context with evidence sampled from another entity pair that shared no entity names with the target pair, to evaluate whether the model could disregard off-target knowledge.

\textbf{Corrupted relation labels.} We replaced the relation labels in the retrieved KB context with labels from the instance's candidate set, excluding the gold label and labels already present in the context, to evaluate the model's sensitivity to altered relational evidence.

Table~\ref{tab:semrep_kb_robustness} reports performance under original and perturbed KB context, without demonstration retrieval or verification. Relative to the original KB context, corrupted relation label decreases micro-F$_1$ from 0.655 to 0.643 and macro-F$_1$ from 0.568 to 0.565. Irrelevant context yields higher micro-F$_1$ (0.662) and macro-F$_1$ (0.583). One possible explanation is that evidence referring to other entities is easier to disregard, whereas original KB evidence may concern the target entities without supporting the relation expressed in the sentence, potentially misleading the model. However, the observed increase does not establish that irrelevant evidence is beneficial or consistently ignored. Future work could investigate whether selecting KB context more closely aligned with the sentence-level context improves performance.

\begin{table*}[t]
\centering
\small
\caption{SemRepGS performance under original and perturbed KB context. The \textsc{norel} class is excluded from micro- and macro-averaging.}
\begin{tabular}{lcccccc}
\toprule
& \multicolumn{3}{c}{Micro} & \multicolumn{3}{c}{Macro} \\
\cmidrule(lr){2-4}\cmidrule(lr){5-7}
Configuration & Precision & Recall & F$_1$ & Precision & Recall & F$_1$ \\
\midrule
\textsc{Baseline} & 0.534 & 0.844 & 0.654 & 0.473 & 0.793 & 0.573 \\
+ Original KB & 0.534 & 0.846 & 0.655 & 0.470 & 0.790 & 0.568 \\
+ Relation-label-corrupted KB & 0.524 & 0.830 & 0.643 & 0.470 & 0.782 & 0.565 \\
+ Irrelevant KB & \textbf{0.541} & \textbf{0.851} & \textbf{0.662} & \textbf{0.483} & \textbf{0.809} & \textbf{0.583} \\
\bottomrule
\end{tabular}
\label{tab:semrep_kb_robustness}
\end{table*}

% \end{thebibliography}
\end{document}